\documentclass[11pt]{article}

\usepackage[preprint]{acl}

\usepackage{times}
\usepackage{latexsym}
\usepackage[T1]{fontenc}
\usepackage[utf8]{inputenc}
\usepackage{microtype}
\usepackage{inconsolata}
\usepackage{graphicx}
\usepackage{booktabs}
\usepackage{multirow}
\usepackage{amsmath}
\usepackage{amssymb}
\usepackage[dvipsnames]{xcolor}
\usepackage{enumitem}
\usepackage{tcolorbox}
\tcbuselibrary{skins, breakable}
\usepackage{amsmath}

\definecolor{agentblue}{HTML}{8FB4DC}
\definecolor{useryellow}{HTML}{FFDD8E}
\definecolor{envteal}{HTML}{70CDBE}
\definecolor{textblue}{HTML}{7AC3DF}
\definecolor{uiorange}{HTML}{F5AA61}
\definecolor{hybridpurple}{HTML}{AC99D2}
\definecolor{costred}{HTML}{EB7E60}

\newtcolorbox{promptbox}[2][gray]{
  colback=#1!6,
  colframe=#1!55,
  coltitle=white,
  fonttitle=\bfseries\small,
  title={#2},
  before skip=8pt,
  after skip=8pt,
  left=10pt,
  right=10pt,
  top=8pt,
  bottom=8pt,
  breakable,
  enhanced,
  sharp corners,
  boxrule=0.7pt,
}

\usepackage{float}
\usepackage[linesnumbered,ruled]{algorithm2e}

\title{Communication Policy Evolution for Proactive LLM Agents}

\author{Xinbei Ma$^{1}$\thanks{Equal contribution. Work done during internship at OPPO.}, Jiyang Qiu$^{1,*}$, Yao Yao$^{1}$, Zheng Wu$^{1}$, Yijie Lu$^{1}$,\\ \textbf{Xiangmou Qu$^{2}$, Jiaxin Yin$^{2}$, Xingyu Lou$^{2, \dagger}$, Jun Wang$^{2, \dagger}$,} \\ \textbf{Weiwen Liu$^{1}$, Weinan Zhang$^{1}$, Zhuosheng Zhang$^{1,\dagger}$, Hai Zhao$^{1}$\thanks{Corresponding authors.}} \\
$^1$Shanghai Jiao Tong University, $^2$OPPO Research Institute\\ 
\texttt{\{sjtumaxb, qiujiyang, zhangzs\}@sjtu.edu.cn},
\texttt{zhaohai@cs.sjtu.edu.cn}\\
\texttt{louxingyu@oppo.com}, \texttt{junwang.lu@gmail.com}, 
}

\begin{document}
\maketitle

\begin{abstract}
LLM agents have rapidly evolved into autonomous systems, yet a persistent information gap remains between users and agents: communication is costly, while users' identical preferences further limit information exchange. 
To investigate \emph{how} agents should communicate across modalities, this paper formalizes \emph{Communication Policy}, establishes textual and UI-based policies, and then evaluates communication policies across diverse environments, personas, and model combinations.
Building information asymmetry for proactive agents, we set up two complementary settings, \emph{User--Agent} and \emph{Planner--Executor}.
Experimental results reveal complementary strengths between interaction channels: text-based interaction often facilitates task performance, while structured UI improves agents' response quality and persona compliance. 
Motivated by that, a hybrid method combines these advantages.
we further propose \emph{Communication Policy Evolution} (CPE), a self-evolution framework for refining communication policies through rollout and prompt-level evolving. Without model modification, CPE achieves the best task success across multiple settings using prompt refinement alone. Our findings identify communication behavior as a critical yet underexplored design dimension for LLM agents.

\end{abstract}

\section{Introduction}

Large language model (LLM) agents have rapidly evolved into autonomous systems capable of reasoning, tool use, and extended interaction with users and environments~\citep{yao2022react,wei2022chain,patil2024gorilla,qin2024toolllm,wang2025human}. Yet despite these advances, a fundamental bottleneck remains: users often hold the complete task they want to achieve, but natural language cannot fully convey all constraints, preferences, and edge cases at once. Important information emerges only gradually through interaction. 
Therefore, agents always have access to 
\textbf{partial information}, where successful task completion depends not only on reasoning and execution, but also on the agent’s ability to recover missing information through interaction~\citep{fang2025information}. The challenge is further compounded by differences in user expertise, patience, and communication style, all of which directly shape interaction quality~\citep{bhattacharjee2024understanding,wu2024navigating}.

Proactive agents actively acquire missing information through clarification dialogues~\citep{deng2023prompting,liao2023proactive}, proactive planning~\citep{zhang2024ask}, and sequential decision-making under uncertainty~\citep{suri2025structured,huang2025teaching}. However, this line of work focuses primarily on \emph{what} the agent should ask, while largely overlooking an equally important question: \emph{how should the agent communicate?} In practice, information recovery is fundamentally shaped by the communication channel itself~\citep{sachdeva2024taking}. Free-form natural language enables flexible and open-ended interaction, while structured interfaces constrain user responses into organized and lower-ambiguity formats. Choosing between these interaction channels therefore becomes a critical decision for effective task-state alignment.

Generative UI uses LLMs to produce HTML and associated code, which is then rendered as multimodal interfaces for users to view and interact with. Recent work shows that structured interfaces can substantially improve information collection quality~\citep{chen2025generative}: compared to unconstrained free-text interaction, structured forms shift users from recall to recognition, improving precision through input constraints and visual organization~\citep{wei2024leveraging,anbalagan2025bridging}. Moreover, HTML interfaces can now be generated fluently by LLMs themselves~\citep{cao2025generative,nandy2024bespoke}, and Agent--Computer Interface research further highlights the value of structured interaction surfaces for LLM agents~\citep{jimenez2024swe}. However, existing work primarily aims to improve Generative UI itself rather than deploy it in real-world applications.


In this paper, we formalize the channel-selection decision as \textbf{Communication Policy}. We equip agents with two communication primitives: {\texttt{ask\_question}} for free-form language interaction and {\texttt{generate\_ui}} for HTML-based forms. This defines two single-channel settings, $M_{\mathrm{text}}$ and $M_{\mathrm{ui}}$, and a hybrid setting, $M_{\mathrm{hybrid}}$, where the agent dynamically selects between the two channels.
Across four environments and diverse user personas, we find that text and UI have complementary strengths: text drives task completion, while UI improves response quality and persona compliance. Hybrid access achieves the best overall results in most settings, though the optimal strategy depends on task structure and persona.
To optimize communication policy automatically, we propose \underline{C}ommunication \underline{P}olicy \underline{E}volution (CPE). In each round, CPE evaluates the current policy on a training batch, prompts an LLM to analyze the rollout results and propose targeted edits, and accepts or rejects the candidate via a two-stage gate that guarantees monotonic improvement on held-out data. The optimized policies achieve best task completion across all evaluated configurations, using only prompt-level refinement without modifying model weights.

Our contributions are three-fold:
\begin{enumerate}[nosep]
\item We identify channel selection as a fundamental problem in LLM agent interaction and formalize Communication Policy for hybrid text+UI communication.

\item We systematically evaluate hybrid communication under partial information, showing that text and UI interaction exhibit complementary strengths across tasks and users.

\item 
We propose CPE, a self-evolution framework that discovers effective communication policies through iterative rollout analysis, achieving best productivity via prompt optimization.

\end{enumerate}

\section{Related Work}

\paragraph{LLM Agents' Proactivity}

Proactive LLM agents actively seek missing information rather than passively following incomplete instructions. Work in this area spans prompting LLMs for clarification dialogues~\citep{deng2023prompting}, structured clarification in dialogue systems~\citep{sahay2025ask,siro2026agent}, proactive planning where agents ask before acting~\citep{zhang2024ask}, anticipating user needs from environmental events~\citep{lu2025proactive}, sequential decision-making under uncertainty~\citep{suri2025structured,huang2025teaching}, and open-ended proactive assistance~\citep{abbas2026having}. Despite this diversity, these efforts focus on \emph{what} to ask: the communication channel itself, \emph{how} the agent asks, remains unexamined. 

\paragraph{User-centric Agent}
Designs of user-centric agents focus on how agents adapt to individual user preferences instead of interacting with all users in the same way. Persona-conditioned user simulators embed diverse interaction styles into LLM-based user proxies, enabling controlled evaluation without costly human studies~\citep{dou2025simulatorarena,gromada2025evaluating,samuel2024personagym,wang2025know}, and power benchmarks probing personalization and human--agent conversational gaps~\citep{hao2025evaluating,wang2025human}. Personalization techniques leverage explicit user profiles, latent preference models, or curiosity-driven rewards to tailor responses~\citep{li2024personalized,qiu2025latent,shi2025personax,wan2026enhancing}. Building on this line of work, we introduce the communication channel itself as a design variable, investigating whether structured UI-based interaction can improve an agent's ability to align with user personas.

\begin{figure*}[t]
\centering
\includegraphics[width=\textwidth]{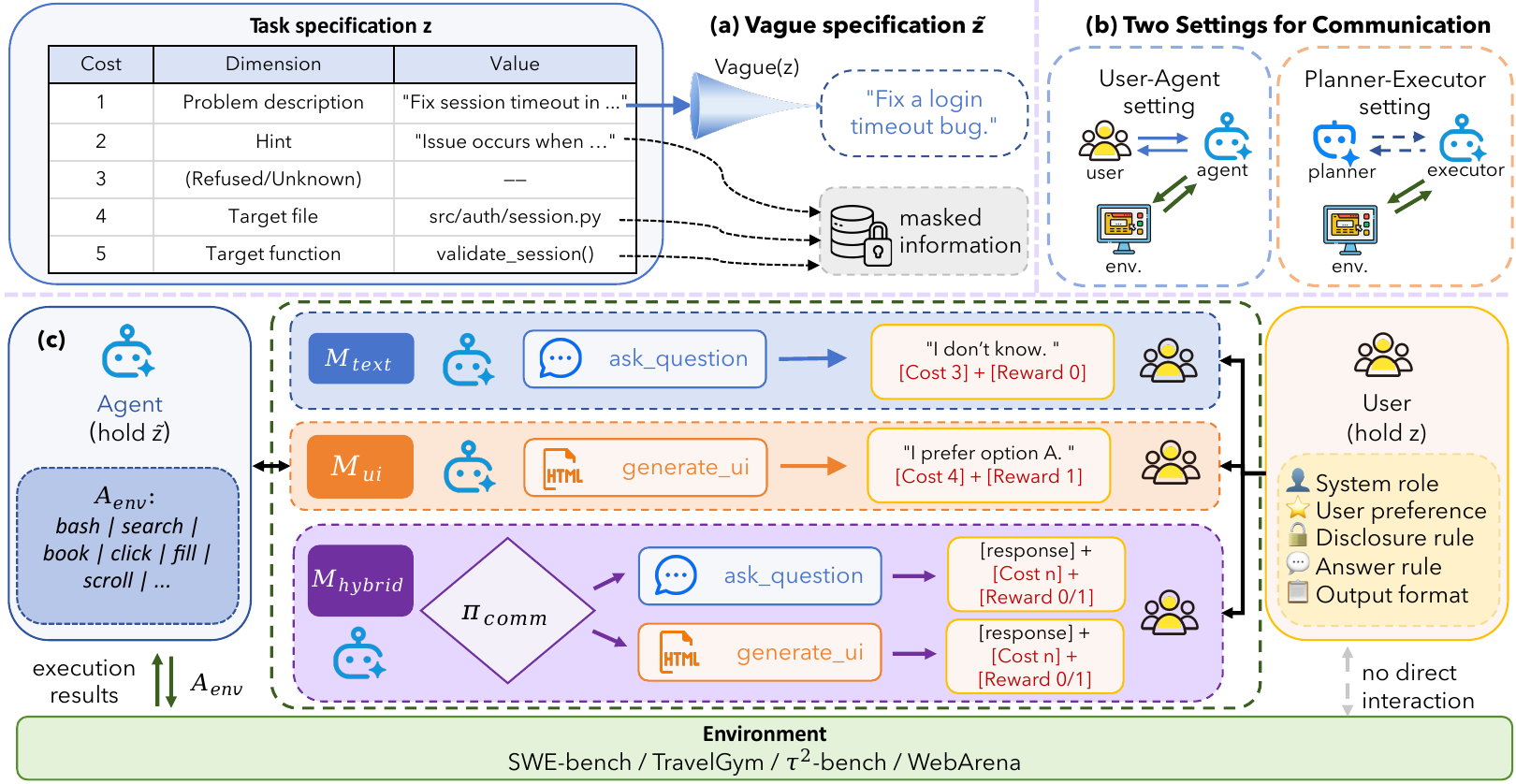}
\caption{Overview of our communication policy formulation and evolution. 
\textbf{(a)} A full task specification $z$ contains information dimensions with different sensitivity costs, while the agent observes only a vague version $\tilde{z}=\text{vague}(z)$. 
\textbf{(b)} We study two settings: User--Agent interaction and Planner--Executor interaction, where the agent/executor interacts with both a simulator and the environment. 
\textbf{(c)} We compare text-only $M_{\text{ui}}$, UI-only $M_{\text{ui}}$, and hybrid communication modes $M_{\text{hybrid}}$, selecting channels via $\pi_{\text{comm}}$ at each turn. 
The simulator records the disclosed Cost level and optional persona-alignment Reward.}
\label{fig:framework}
\end{figure*}

\paragraph{Test-time Evolving}
Recent test-time optimization methods enable LLM agents to improve behavior through interaction experience without parameter updates. Existing approaches include black-box prompt optimization~\citep{yang2024large,wang2024promptagent}, textual gradient descent~\citep{yuksekgonul2024textgrad}, trace-driven reflection with gating~\citep{agrawal2025gepa,yi2025zera}, and multi-agent optimization~\citep{zhang2026mapro}. Related self-reflective agents update memory from failure traces~\citep{shinn2023reflexion,zhao2024expel}, although their effectiveness often depends on the quality of verification signals~\citep{huang2024large}. While these methods optimize prompts to improve agents' task execution capabilities, our work instead focuses on optimizing communication behavior, training agents to decide when to communicate through natural language and when to rely on structured UI interactions in order to recover user intent under partial information.

\section{Communication Policy Evaluation}

\subsection{Problem Formulation}
\label{sec:partial-info}
We consider an interaction setting with three components: a user, an agent, and an environment.
An achievable task is fully specified by a \emph{task specification} $z$, which contains all information necessary for successful execution and is held by the proposer, e.g., the user.
Due to information loss during communication, the agent only observes a \emph{vague specification} $\tilde{z}$, in which part of the task-relevant information in $z$ is missing or underspecified.
\begin{equation}
\tilde{z} = \text{vague}(z), \quad \text{where } |\tilde{z}| \ll |z|.
\label{eq:vague-spec}
\end{equation}




The interaction proceeds in steps $t = 1, 2, \ldots$. At step $t$, the agent holds a belief $b_t$ about $z$, initialized from $\tilde{z}$, and selects a communication action $a_t$ from the action space determined by the interaction mode (\S\ref{sec:comm-modes}). The user sees $a_t$ together with the interaction history $h_{<t} = (a_1, d_1, \ldots, a_{t-1}, d_{t-1})$, and produces a disclosure:
\begin{equation}
d_t \sim \mathcal{M}(z, a_t, h_{<t}),
\label{eq:disclosure}
\end{equation}
where $\mathcal{M}$ is the user's response policy. The agent updates its belief:
\begin{equation}
b_{t+1} = \text{Update}(b_t, d_t).
\label{eq:belief-update}
\end{equation}

The interaction ends when the agent issues a task-execution action. The agent's objective is to choose communication actions that maximize task success while keeping the disclosure cost in check:
\begin{equation}
\max_{\{a_t\}_{t=1}^{T}} \; \mathcal{T}(o_T; z) \quad \text{s.t.} \quad \sum_{t=1}^{T} c_t \leq C_{\max},
\label{eq:agent-goal}
\end{equation}
where $T$ is the terminal step, $o_T$ is the output produced from the final belief $b_T$, $\mathcal{T}(\cdot; z)$ is an environment-specific success function (e.g., patch correctness, terminal reward, database-state match), and $c_t$ is the sensitivity cost of disclosure $d_t$. 
Effective communication in this setting requires agents to complete tasks successfully (\emph{productivity}), proactively recover missing information (\emph{proactivity}), and adapt to user-specific interaction preferences (\emph{personalization}) \cite{sun2025training}.

\subsection{Communication Modes}
\label{sec:comm-modes}

We first introduce two communication primitives beyond the task actions $\mathcal{A}_{\text{env}}$.
The primitive \texttt{ask\_question} sends a natural-language query and receives a free-text response, while \texttt{generate\_ui} renders an HTML form as a screenshot for the user to complete.
Based on these primitives, we define two single-channel modes and one hybrid mode:
\begin{equation}
\begin{aligned}
\mathcal{A}_{\text{text}} &= \mathcal{A}_{\text{env}} \cup \{\texttt{ask\_question}\}, \\
\mathcal{A}_{\text{ui}}   &= \mathcal{A}_{\text{env}} \cup \{\texttt{generate\_ui}\}, \\
\mathcal{A}_{\text{hybrid}} &= \mathcal{A}_{\text{env}} \cup \{\texttt{ask\_question}, \texttt{generate\_ui}\}.
\end{aligned}
\label{eq:modes}
\end{equation}

We denote the resulting modes as $M_{\text{text}}$, $M_{\text{ui}}$, and $M_{\text{hybrid}}$, respectively.
$M_{\text{text}}$ and $M_{\text{ui}}$ are single-channel modes: the former supports flexible free-text interaction, while the latter supports structured information collection through fixed fields.
$M_{\text{hybrid}}$ further combines both primitives, allowing the agent to choose between text and UI communication at each turn.
We formalize this channel-selection mechanism as the \emph{Communication Policy}.
Figure~\ref{fig:framework} summarizes the three modes and the interaction loop.

\subsection{Two Interaction Settings}
\label{sec:two-settings}

We design two settings that isolate different sources of the information gap: 

\smallskip

\noindent\textbf{Setting A: User--Agent.}
This is our primary setting, designed to approximate realistic user-facing agent interaction.
The user is simulated by a \textit{User Simulator}, an LLM parameterized by a persona (\S\ref{sec:personas}).
The user holds the full task specification $z$, while the agent receives only the vague specification $\tilde{z}$ and must recover missing information through communication.
The persona captures user-specific interaction preferences and shapes disclosure behavior, making responses potentially uncertain, incomplete, indirect, or evolving over time.
Accordingly, this setting evaluates task completion, proactive recovery of missing information, and adaptation to user preferences.
The full \textit{User Simulator} system prompt is detailed in \S\ref{sec:prompt-templates}.

\smallskip
\noindent\textbf{Setting B: Planner--Executor.}
This auxiliary setting abstracts away subjective user preferences and focuses on agent--agent collaboration.
The \textit{Planner Simulator} holds the full task specification $z$ and provides global planning guidance, while the executor receives only $\tilde{z}$ and must query the planner to recover missing execution details.
Since both roles are agents, no persona is introduced; the planner discloses information objectively under the cost schema.
Thus, this setting contains an information gap between planner and executor, but removes personalization-related factors and evaluates only collaborative task completion.
The full \textit{Planner Simulator} system prompt is detailed in \S\ref{sec:planner-prompts}.

\section{Communication Policy Evolution}

\subsection{The Communication Policy}

While $M_{\text{hybrid}}$ provides access to both communication channels and shows strong potential, it does not specify how the channels should be used.
To further exploit the hybrid setting, we propose \textbf{Communication Policy Evolution} (CPE), a training-free prompt evolving method, which optimizes this policy to guide when the agent should use text or UI interaction.

$$
\pi_{\text{comm}} = (\text{system\_prompt}, \; \text{examples}, \; \text{appendixes}).
$$

The system prompt specifies the available tools, their invocation formats, and heuristics for channel selection. The examples provide few-shot demonstrations of choosing between \texttt{ask\_question} and \texttt{generate\_ui} under varied conditions. The appendixes carry environment-specific guidance. The policy is rendered to the agent at the start of each episode; optimizing it reduces to rewriting text informed by rollout results. Example prompts for all four benchmarks are provided in Appendix~\ref{sec:comm-policy-prompts}.

\begin{figure}[t]
\centering
\includegraphics[width=\columnwidth]{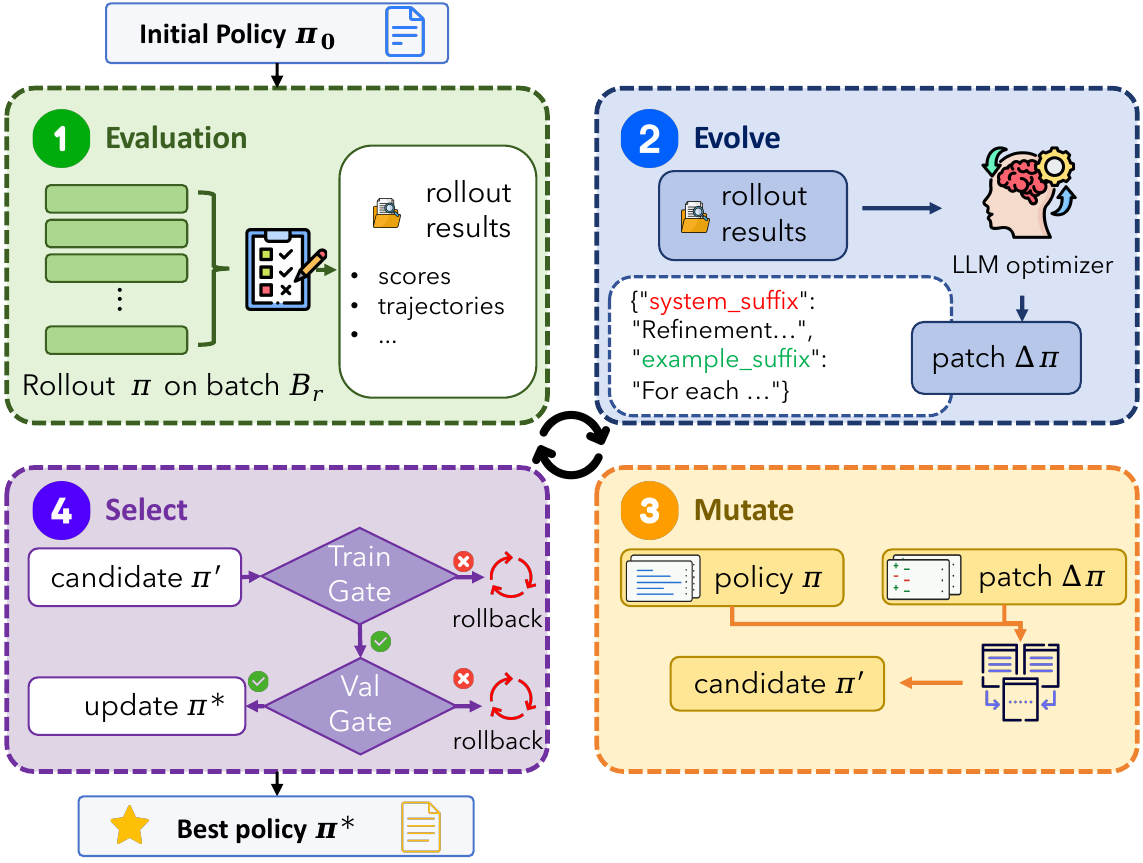}
\caption{Communication Policy Evolution (CPE). Each round: (1) \textbf{Evaluate} the current policy $\pi$ on a batch $\mathcal{B}_r$; (2) \textbf{Evolve}: the LLM analyzes hybrid-only signals (scores, trajectories, task specs, current policy, patch history) and proposes a JSON patch $\Delta\pi$; (3) \textbf{Mutate} by applying $\Delta\pi$ as text overrides to produce candidate $\pi'$; (4) \textbf{Select} via two-stage gating: candidate must beat $\pi$ on $\mathcal{B}_r$ (train accept), then beat $\pi^*$ on $\mathcal{D}_{\text{val}}$ (val accept). Val-accepted candidates update $\pi^*$, guaranteeing monotonic improvement.}
\label{fig:cpe-loop}
\end{figure}

\subsection{Policy Optimization for Communication Policy Evolution}
\label{sec:cpe}

CPE (Figure~\ref{fig:cpe-loop}) evolves the communication policy $\pi_{\text{comm}}$ by iterative refinement against rollout results. 
Given a task distribution $\mathcal{D}$, an initial communication policy $\pi_0$, and a total round budget $R$, CPE performs prompt-level optimization over $\pi_{\text{comm}}$.
It rewrites only the communication-policy prompt, while leaving the underlying agent and user-simulator model weights unchanged.

\smallskip
\noindent\textbf{Objective and regret.}
For a task-execution agent, completing the task remains the primary goal. We therefore optimize for productivity, defining the CPE objective as the mean productivity score over the task distribution:
\begin{equation}
J(\pi_{\text{comm}}; \mathcal{D}) = \frac{1}{|\mathcal{D}|} \sum_{i \in \mathcal{D}} s_{\text{prod}}^{(i)}.
\label{eq:J}
\end{equation}

Proactivity and personalization are reported throughout to make the trade-off visible: when communication policy is optimized for productivity, both interaction scores may rise (better channel choices elicit higher-quality responses) or dip (the policy trades interaction turns for execution speed). We treat this tension as a descriptive finding rather than a multi-objective optimization problem.

Giving the agent both channels does not tell it how to use them. When hybrid underperforms a single-channel baseline, the agent is worse off than if it had been restricted to one channel. We define the \textbf{instance-level regret} to quantify this gap:
\begin{equation}
r^{(i)} = \max\bigl(s_{\text{text}}^{(i)},\, s_{\text{ui}}^{(i)}\bigr)
- s_{\text{hybrid}}^{(i)},
\label{eq:regret}
\end{equation}
where $r^{(i)} > 0$ means hybrid underperformed a single-channel baseline, and minimizing regret is equivalent to maximizing $J(\pi_{\text{comm}}; \mathcal{D})$.

\smallskip
\noindent\textbf{Iterative refinement.}
Each round of CPE proceeds in four steps:

\begin{enumerate}[nosep,leftmargin=2em]
    \item \textbf{Evaluate.} The current policy $\pi_{\text{comm}}$ is rolled out on a batch $\mathcal{B} \subseteq \mathcal{D}_{\text{train}}$ of $K$ episodes, producing per-episode scores and interaction trajectories $\tau^{(i)} = \{(a_t, o_t, c_t)\}_{t=1}^{T_i}$, where $a_t \in \{\texttt{ask\_question}, \texttt{generate\_ui}\}$ is the agent's channel choice at turn $t$, $o_t$ is the user's response, and $c_t$ is the Cost incurred (Eq.~\ref{eq:agent-goal}).

    \item \textbf{Evolve.} The same LLM is prompted to self-evolve its communication policy: it analyzes the batch evaluation results and proposes targeted edits. The signal provided to the LLM is detailed in \S\ref{sec:evolve-evidence}. From these signals, the LLM produces a structured JSON patch:
    \begin{equation}
    \begin{aligned}
    \Delta\pi_{\text{comm}} = \text{Evolve}\bigl(
    &\pi_{\text{comm}}, \\
    &\{s_{\text{prod}}^{(i)}, s_{\text{pro}}^{(i)}, s_{\text{pers}}^{(i)}\}_{i \in \mathcal{B}}, \\
    &\{\tau^{(i)}, z^{(i)}\}_{i \in \mathcal{B}},\;
    \mathcal{H} \bigr),
    \end{aligned}
    \label{eq:evolve}
    \end{equation}
where $\Delta\pi_{\text{comm}}$ is a JSON patch applied over the current policy text, overwriting or appending to its components.

    \item \textbf{Mutate.} The patch is applied to produce a candidate policy $\pi_{\text{comm}}'$.

    \item \textbf{Select.} $\pi_{\text{comm}}'$ is evaluated on the same batch $\mathcal{B}$. The candidate is accepted if:
    \begin{equation}
    J(\pi_{\text{comm}}'; \mathcal{B}) > J(\pi_{\text{comm}}; \mathcal{B}) + \epsilon.
    \label{eq:accept-train}
    \end{equation}
    If rejected, the overrides are rolled back to the pre-round state and the policy remains $\pi_{\text{comm}}$.
\end{enumerate}

\smallskip
\noindent\textbf{Validation gating and monotonicity.}
To prevent overfitting to individual batches, $\mathcal{D}$ is partitioned into $\mathcal{D}_{\text{train}}$ and $\mathcal{D}_{\text{val}}$, and a best-so-far policy $\pi_{\text{comm}}^*$ with score $J^*$ is maintained across rounds. Batches $\mathcal{B}_r \subseteq \mathcal{D}_{\text{train}}$ of size $K$ are drawn by epoch-style sampling without replacement: the pool is shuffled, contiguous blocks are drawn sequentially, and the pool is reshuffled upon exhaustion. When a candidate passes the train accept condition (Eq.~\ref{eq:accept-train}), it is evaluated on the full $\mathcal{D}_{\text{val}}$:
\begin{equation}
J(\pi_{\text{comm}}'; \mathcal{D}_{\text{val}}) > J^* + \epsilon.
\label{eq:accept-val}
\end{equation}

Only when this holds is $\pi_{\text{comm}}^*$ replaced by $\pi_{\text{comm}}'$ and $J^*$ updated. By construction, $J^*$ is \textbf{monotonically non-decreasing} over rounds: the best policy on held-out data never degrades. The working policy $\pi_{\text{comm}}$ may fluctuate as it tracks batch-level variation, but $\pi_{\text{comm}}^*$ records the strongest generalizing policy encountered. Figure~\ref{fig:cpe-loop} illustrates the procedure; the formal algorithm is provided in algorithm~\ref{alg:cpe}.

\begin{algorithm}[t]
\small
\SetAlgoSkip{small}
\KwIn{Task distribution $\mathcal{D}$, initial policy $\pi_0$, rounds $R$, batch size $K$, tolerance $\epsilon$}
\KwOut{Best policy $\pi^*$}
$\pi \leftarrow \pi_0$; $\pi^* \leftarrow \pi_0$; $J^* \leftarrow J(\pi_0; \mathcal{D}_{\text{val}})$; $\mathcal{H} \leftarrow \emptyset$\;
\For{$r = 1$ \KwTo $R$}{
    $\mathcal{B}_r \leftarrow \text{SampleEpoch}(\mathcal{D}_{\text{train}}, K)$\;
    $J_{\text{pre}} \leftarrow J(\pi; \mathcal{B}_r)$ \tcp*{1. Evaluate}
    $\Delta\pi \leftarrow \text{Evolve}(\pi, \text{scores}(\mathcal{B}_r), \text{trajectories}(\mathcal{B}_r), \mathcal{H})$ \tcp*{2. Evolve}
    $\pi' \leftarrow \text{Apply}(\pi, \Delta\pi)$ \tcp*{3. Mutate}
    $J_{\text{post}} \leftarrow J(\pi'; \mathcal{B}_r)$ \tcp*{4. Select}
    \If{$J_{\text{post}} > J_{\text{pre}} + \epsilon$}{
        $\pi \leftarrow \pi'$\;
        $J_{\text{val}} \leftarrow J(\pi'; \mathcal{D}_{\text{val}})$\;
        $\mathcal{H} \leftarrow \mathcal{H} \cup \{(\Delta\pi, J_{\text{val}}, J_{\text{val}} > J^* + \epsilon)\}$\;
        \If{$J_{\text{val}} > J^* + \epsilon$}{
            $\pi^* \leftarrow \pi'$, $J^* \leftarrow J_{\text{val}}$\;
        }
    }
}
\Return{$\pi^*$}\;
\caption{Communication Policy Evolution (CPE)}
\label{alg:cpe}
\end{algorithm}

\subsection{Evolution Signals}
\label{sec:evolve-evidence}

The reflect LLM receives five categories of signals, all drawn from $M_{\text{hybrid}}$ rollouts:

\begin{itemize}[nosep,leftmargin=2em]
    \item \textit{Scores.} For each $i \in \mathcal{B}$, all three per-dimension scores $\{s_{\text{prod}}^{(i)}, s_{\text{pro}}^{(i)}, s_{\text{pers}}^{(i)}\}$ under $M_{\text{hybrid}}$, with $s_{\text{prod}}^{(i)}$ as the optimization target.
    \item \textit{Trajectories.} The full rollouts $\{\tau^{(i)}\}$: each agent communication action tagged by channel type (\texttt{ask\_question} or \texttt{generate\_ui}), its content, and the user's response including Cost annotations.
    \item \textit{Task specifications.} The system messages sent to the user, containing the ground-truth $z$.
    \item \textit{Current policy.} The complete text of $\pi_{\text{comm}}$, incorporating all edits accumulated from prior accepted rounds.
    \item \textit{Patch history.} The most recent accepted and rejected patches from prior rounds, each annotated with its validation outcome.
\end{itemize}

\section{Experimental Setup}
\begin{table*}[t]
\centering
\small
\resizebox{0.85\textwidth}{!}{%
\begin{tabular}{lllccc}
\toprule
\textbf{Benchmark} & \textbf{Agent} & \textbf{User} & \textbf{$M_{\text{text}}$} & \textbf{$M_{\text{ui}}$} & \textbf{$M_{\text{hybrid}}$} \\
\midrule
SWE-bench  & Qwen3-32B     & Qwen3-VL-32B & .035/\textcolor{uiorange}{\textbf{.250}}/.355 & .009/.161/.421 & \textcolor{agentblue}{\textbf{.040}}/.134/\textcolor{envteal}{\textbf{.454}} \\
           & Seed-OSS-36B  & Qwen3-VL-32B & .135/.076/\textcolor{envteal}{\textbf{.545}} & .120/\textcolor{uiorange}{\textbf{.156}}/.502 & \textcolor{agentblue}{\textbf{.139}}/.085/.520 \\
           & GPT-5-mini    & Qwen3-VL-32B & .031/.129/.472 & .030/\textcolor{uiorange}{\textbf{.179}}/.589 & \textcolor{agentblue}{\textbf{.043}}/.138/\textcolor{envteal}{\textbf{.667}} \\
           & DeepSeek-V3.2 & Qwen3-VL-32B & .135/.156/.629 & .113/\textcolor{uiorange}{\textbf{.196}}/\textcolor{envteal}{\textbf{.705}} & \textcolor{agentblue}{\textbf{.140}}/.089/.594 \\
\midrule
TravelGym & DeepSeek-V3.2 & GPT-4o        & \textcolor{agentblue}{\textbf{1.437}}/.045/.905 & 1.249/\textcolor{uiorange}{\textbf{.260}}/\textcolor{envteal}{\textbf{.952}} & 1.233/.145/.856 \\
           & GPT-5-mini    & GPT-4o        & 1.055/.250/.695 & .998/\textcolor{uiorange}{\textbf{.470}}/\textcolor{envteal}{\textbf{.890}} & \textcolor{agentblue}{\textbf{1.055}}/.250/.710 \\
           & DeepSeek-V3.2 & Qwen3-VL-32B & 1.365/\textcolor{uiorange}{\textbf{.220}}/.895 & 1.416/.135/\textcolor{envteal}{\textbf{.950}} & \textcolor{agentblue}{\textbf{1.514}}/.120/.915 \\
           & GPT-5-mini    & Qwen3-VL-32B & \textcolor{agentblue}{\textbf{1.013}}/\textcolor{uiorange}{\textbf{.345}}/.850 & .995/.285/\textcolor{envteal}{\textbf{.865}} & .906/.195/.766 \\
\midrule
$\tau^2$-bench & DeepSeek-V3.2 & GPT-4o        & .270/\textcolor{uiorange}{\textbf{.230}}/.650 & \textcolor{agentblue}{\textbf{.275}}/.120/\textcolor{envteal}{\textbf{.850}} & .270/.015/.610 \\
           & DeepSeek-V3.2 & Qwen3-VL-32B & .280/.020/.485 & .235/\textcolor{uiorange}{\textbf{.035}}/\textcolor{envteal}{\textbf{.555}} & \textcolor{agentblue}{\textbf{.300}}/\textcolor{uiorange}{\textbf{.035}}/.495 \\
           & GPT-5-mini    & Qwen3-VL-32B & \textcolor{agentblue}{\textbf{.375}}/.015/.347 & \textcolor{agentblue}{\textbf{.375}}/.105/\textcolor{envteal}{\textbf{.593}} & .340/\textcolor{uiorange}{\textbf{.135}}/.470 \\
           & GPT-5-mini    & GPT-4o        & \textcolor{agentblue}{\textbf{.425}}/.045/.643 & .385/\textcolor{uiorange}{\textbf{.065}}/\textcolor{envteal}{\textbf{.827}} & .405/.075/.563 \\
\midrule
WebArena  & DeepSeek-V3.2 & GPT-4o        & \textcolor{agentblue}{\textbf{.205}}/.628/\textcolor{envteal}{\textbf{.865}} & .180/\textcolor{uiorange}{\textbf{.810}}/.673 & .195/.695/.766 \\
           & GPT-5-mini    & GPT-4o        & \textcolor{agentblue}{\textbf{.200}}/.525/.737 & .140/\textcolor{uiorange}{\textbf{.668}}/\textcolor{envteal}{\textbf{.803}} & .180/.315/.752 \\
\bottomrule
\end{tabular}%
}
\caption{Mode comparison of $M_{\text{text}}$, $M_{\text{ui}}$, and $M_{\text{hybrid}}$ across four benchmarks in the User--Agent setting. Each cell reports productivity / proactivity / personalization, averaged over four personas. Best value per metric per row is color-coded: \textcolor{agentblue}{productivity}, \textcolor{uiorange}{proactivity}, \textcolor{envteal}{personalization}.}
\label{tab:main-mode-compare}
\end{table*}

\subsection{Environments}

We evaluate across four environments spanning diverse domains and interaction settings: SWE-bench (software repair)~\citep{jimenez2024swe}, TravelGym (travel planning)~\citep{qian2025userbench}, $\tau^2$-bench (customer service)~\citep{barres2025tau}, and WebArena (web navigation)~\citep{zhou2024webarena}. Together, these environments cover code editing, preference elicitation, conversational task completion, and browser-based interaction under partial information.
Detailed environment specifications, task formulations, and the exact mapping from $z$ to $\tilde{z}$ are provided in Appendix~\ref{sec:env-details} (Table~\ref{tab:envs}).






\subsection{User and Personas}

In Setting~A (User--Agent), the user is simulated by an LLM that holds the full task specification $z$ (per-benchmark schemas in Appendix~\ref{sec:env-details} and~\ref{sec:cost-defs}), is assigned one of four personas (Appendix~\ref{sec:personas}), and replies to agent queries with Cost annotations following the disclosure rule of Eq.~\ref{eq:disclosure}. Each disclosure $d_t$ is annotated with a Cost label indicating the highest sensitivity level revealed; the user follows a minimal-disclosure principle, revealing only the lowest-cost information sufficient to answer the query and refusing when no matching information is available.
In Setting~B (Planner--Executor, \S\ref{sec:two-settings}), the user is replaced by a planner LLM holding the same information. The planner has no persona.
\subsection{Interaction Modes and Models}

We evaluate all three communication modes defined in \S\ref{sec:comm-modes}: $M_{\text{text}}$, $M_{\text{ui}}$, and $M_{\text{hybrid}}$. Agent models are DeepSeek-V3.2~\citep{liu2025deepseek}, GPT-5-mini, Qwen3-VL-32B-Instruct~\citep{bai2025qwen3}, and Seed-OSS-36B-Instruct~\citep{seed2025seed-oss}; user-side models are GPT-4o~\citep{achiam2023gpt} and Qwen3-VL-32B-Instruct. Specific pairings vary by benchmark and are reported in each result table.

\subsection{Metrics}
\label{sec:metrics}

Following \citet{sun2025training} and the facets of communication quality in \S\ref{sec:comm-modes}, each episode yields three scores:

\textbf{Productivity.} Whether the agent completes the task successfully, measured by each environment's native task-success signal (e.g., patch similarity, database-state match), with TravelGym max~2.4. This is the primary signal: communication is valuable only insofar as it enables task completion.

\textbf{Proactivity.} To complete the task, the agent must act in the environment while recovering the missing information through the communication channel.
Each piece of information in $z$ carries a \emph{sensitivity level} $\ell \in \{1, \ldots, L\}$.
The inverse of interaction cost, computed from the Cost annotations tagged by the \textit{User Simulator} at each turn: a proactive agent achieves high productivity with low cumulative cost $\sum_t c_t$ and few questioning turns.

\textbf{Personalization.} Adherence to the assigned persona's interaction preference. The \textit{User Simulator} tags each response with a persona-specific reward label (0/1); the score is the negative of the cumulative penalty for deviations. In Setting~B, personas are not used and this dimension is absent.

\section{Experiments and Analysis}
\begin{table*}[t]
\centering
\resizebox{\textwidth}{!}{%
\begin{tabular}{lllcccc}
\toprule
\textbf{Benchmark} & \textbf{Agent} & \textbf{User} & \textbf{$M_{\text{text}}$} & \textbf{$M_{\text{ui}}$} & \textbf{$M_{\text{hybrid}}$} & \textbf{$M_{\text{CPE}}$} \\
\midrule
SWE-bench  & DeepSeek-V3.2 & Qwen3-VL-32B & .135/.156/.629 & .113/.196/.705 & .140/.089/.594 & \textcolor{agentblue}{\textbf{.214}}\textsuperscript{$\uparrow$}/.076\textsuperscript{$\downarrow$}/.589\textsuperscript{$\downarrow$} \\
\midrule
$\tau^2$-bench & DeepSeek-V3.2 & GPT-4o        & .270/.230/.650 & .275/.120/.850 & .270/.015/.610 & \textcolor{agentblue}{\textbf{.300}}\textsuperscript{$\uparrow$}/.005\textsuperscript{$\downarrow$}/.520\textsuperscript{$\downarrow$} \\
           & DeepSeek-V3.2 & Qwen3-VL-32B  & .280/.020/.485 & .235/.035/.555 & .300/.035/.495 & \textcolor{agentblue}{\textbf{.300}}\textsuperscript{$-$}/.055\textsuperscript{$\uparrow$}/.575\textsuperscript{$\uparrow$} \\
           & GPT-5-mini    & Qwen3-VL-32B  & .375/.015/.347 & .375/.105/.593 & .340/.135/.470 & \textcolor{agentblue}{\textbf{.380}}\textsuperscript{$\uparrow$}/.145\textsuperscript{$\uparrow$}/.603\textsuperscript{$\uparrow$} \\
           & GPT-5-mini    & GPT-4o        & .425/.045/.643 & .385/.065/.827 & .405/.075/.563 & \textcolor{agentblue}{\textbf{.430}}\textsuperscript{$\uparrow$}/.075\textsuperscript{$-$}/.630\textsuperscript{$\uparrow$} \\
\midrule
TravelGym & DeepSeek-V3.2 & GPT-4o        & 1.437/.045/.905 & 1.249/.260/.952 & 1.233/.145/.856 & \textcolor{agentblue}{\textbf{1.533}}\textsuperscript{$\uparrow$}/.170\textsuperscript{$\uparrow$}/.920\textsuperscript{$\uparrow$} \\
           & GPT-5-mini    & Qwen3-VL-32B  & 1.013/.345/.850 & .995/.285/.865 & .906/.195/.766 & \textcolor{agentblue}{\textbf{1.113}}\textsuperscript{$\uparrow$}/.205\textsuperscript{$\uparrow$}/.845\textsuperscript{$\uparrow$} \\
\midrule
WebArena  & DeepSeek-V3.2 & GPT-4o        & .205/.628/.865 & .180/.810/.673 & .195/.695/.766 & \textcolor{agentblue}{\textbf{.235}}\textsuperscript{$\uparrow$}/.775\textsuperscript{$\uparrow$}/.633\textsuperscript{$\downarrow$} \\
           & GPT-5-mini    & GPT-4o        & .200/.525/.737 & .140/.668/.803 & .180/.315/.752 & \textcolor{agentblue}{\textbf{.275}}\textsuperscript{$\uparrow$}/.520\textsuperscript{$\uparrow$}/.835\textsuperscript{$\uparrow$} \\
\bottomrule
\end{tabular}%
}
\caption{CPE-optimized policy ($M_{\text{CPE}}$) vs.\ baselines. CPE is applied only where $M_{\text{hybrid}}$ underperforms at least one single-channel baseline in productivity. Best \textcolor{agentblue}{productivity} per row is bolded. $\uparrow$  / $\downarrow$ / $-$ indicate direction of change from $M_{\text{hybrid}}$.}
\label{tab:opt}
\end{table*}

We organize experiments around four analyses: (1) mode comparison ($M_{\text{text}}$ vs.\ $M_{\text{ui}}$ vs.\ $M_{\text{hybrid}}$), (2) oracle full-information upper bound on SWE-bench, (3) CPE-optimized policy ($M_{\text{CPE}}$), and (4) Planner--Executor setting.

\subsection{Mode Comparison}

Table~\ref{tab:main-mode-compare} reports the three modes across four benchmarks in the User--Agent setting, averaged over four personas. Several patterns stand out:

\textbf{$M_{\text{hybrid}}$ leads productivity in the majority of cases, but the gain is task-dependent.} $M_{\text{hybrid}}$ achieves the best productivity in 8 of 14 agent--user pairs, confirming that access to both channels generally improves task completion. The benefit is most consistent in SWE-bench (4/4) and absent in WebArena (0/2). Hybrid access is broadly useful but not universally cost-effective.

\textbf{$M_{\text{ui}}$ dominates personalization and proactivity.} $M_{\text{ui}}$ achieves the best personalization in 10 of 14 pairs and the best proactivity in 8 of 14. The effect is strongest in $\tau^2$-bench and TravelGym, where $M_{\text{ui}}$ leads personalization in all 8 pairs. Structured forms reduce response ambiguity by constraining user input to relevant dimensions, which simultaneously improves persona compliance and elicits more information per turn than free-text replies.

\textbf{$M_{\text{text}}$ outperforms $M_{\text{ui}}$ on productivity.} In 9 of 14 pairs, $M_{\text{text}}$ productivity exceeds $M_{\text{ui}}$. Free-text exchange is more efficient for task progress: the agent can ask precisely targeted questions and receive context-rich replies, whereas UI forms impose a fixed per-turn overhead that does not always pay off in task completion. The productivity gap, together with $M_{\text{ui}}$'s advantage in interaction quality, confirms that no single channel dominates, motivating learned channel-selection strategies.

\subsection{Oracle: Value of Full Information}
\label{sec:oracle}

\begin{table}[htbp]
\centering
\resizebox{\columnwidth}{!}{%
\begin{tabular}{llccc}
\toprule
\textbf{Agent} & \textbf{User} & \textbf{$M_{\text{text}}$} & \textbf{$M_{\text{full}}$} & \textbf{ask\_q} \\
 & & prod/pro/pers & prod/pro/pers & $M_{\text{text}} \rightarrow M_{\text{full}}$ \\
\midrule
DeepSeek-V3.2 & Qwen3-VL-32B & .135/.156/.629 & \textcolor{agentblue}{\textbf{.387}}/.174/.696 & 1.00 $\rightarrow$ 0.96 \\
GPT-5-mini    & Qwen3-VL-32B & .031/.129/.472 & \textcolor{agentblue}{\textbf{.329}}/.152/.454 & 0.80 $\rightarrow$ 0.70 \\
Seed-OSS-36B  & Qwen3-VL-32B & .135/.076/.545 & \textcolor{agentblue}{\textbf{.317}}/.134/.679 & 0.99 $\rightarrow$ 0.59 \\
Qwen3-32B     & Qwen3-VL-32B & .035/.250/.355 & \textcolor{agentblue}{\textbf{.285}}/.201/.211 & 0.73 $\rightarrow$ 0.17 \\
\bottomrule
\end{tabular}%
}
\caption{Oracle experiment on SWE-bench. $M_{\text{full}}$ injects the complete task description into the agent's initial prompt. The \textbf{ask\_q} column reports the proportion of tasks where the agent used \texttt{ask\_question} at least once. Best \textcolor{agentblue}{productivity} per row is bolded.}
\label{tab:oracle}
\end{table}

To isolate the role of information availability, we conduct an oracle experiment on SWE-bench where the complete task specification is injected into the agent's initial prompt. This removes the need for information recovery and isolates task-execution capability from interaction quality.

Table~\ref{tab:oracle} reports the results. Full information improves productivity by 2.3$\times$--10.6$\times$ over $M_{\text{text}}$, confirming that \textbf{missing information, rather than execution ability, is the primary bottleneck}. Correspondingly, \texttt{ask\_question} usage drops consistently under the oracle setting, indicating that most interaction turns in $M_{\text{text}}$ are spent recovering task information.


\subsection{CPE-Optimized Policy}

We apply CPE only to settings where naive hybrid interaction underperforms at least one single-channel baseline, indicating ineffective channel selection. Configuration details are in Appendix~\ref{sec:cpe-config}. Table~\ref{tab:opt} reports $M_{\text{CPE}}$ alongside all baselines for completed optimization runs.

\begin{table}[htbp]
\centering
\resizebox{\columnwidth}{!}{%
\begin{tabular}{lllccc}
\toprule
\textbf{Benchmark} & \textbf{Executor} & \textbf{Planner} & \textbf{$M_{\text{text}}$} & \textbf{$M_{\text{ui}}$} & \textbf{$M_{\text{hybrid}}$} \\
\midrule
SWE-bench  & Seed-OSS-36B  & Qwen3-VL-32B & .298 & \textcolor{agentblue}{\textbf{.352}} & .329 \\
           & Qwen3-32B     & Qwen3-VL-32B & .196 & .149 & \textcolor{agentblue}{\textbf{.201}} \\
\midrule
TravelGym  & DeepSeek-V3.2 & GPT-4o        & 1.676 & 1.760 & \textcolor{agentblue}{\textbf{1.792}} \\
           & GPT-5-mini    & GPT-4o        & 1.552 & 1.544 & \textcolor{agentblue}{\textbf{1.608}} \\
           & DeepSeek-V3.2 & Qwen3-VL-32B  & 1.512 & 1.276 & \textcolor{agentblue}{\textbf{1.672}} \\
           & GPT-5-mini    & Qwen3-VL-32B  & 1.348 & 1.616 & \textcolor{agentblue}{\textbf{1.648}} \\
\midrule
$\tau^2$-bench & DeepSeek-V3.2 & GPT-4o        & \textcolor{agentblue}{\textbf{.240}} & .220 & .140 \\
           & DeepSeek-V3.2 & Qwen3-VL-32B  & .320 & \textcolor{agentblue}{\textbf{.340}} & .200 \\
           & GPT-5-mini    & GPT-4o        & .360 & .360 & \textcolor{agentblue}{\textbf{.400}} \\
           & GPT-5-mini    & Qwen3-VL-32B  & .280 & .240 & \textcolor{agentblue}{\textbf{.380}} \\
\bottomrule
\end{tabular}%
}
\caption{Mode comparison in the Planner--Executor setting. Each cell reports productivity; best per row in \textcolor{agentblue}{bold blue}.}
\label{tab:planner-executor}
\end{table}
Across all 9 experiment configurations, $M_{\text{CPE}}$ consistently achieves the best productivity across all evaluated settings using only prompt-level optimization. The arrows in Table~\ref{tab:opt} show that relative to $M_{\text{hybrid}}$, CPE improves proactivity in 6 of 9 cases and personalization in 6 of 9 cases, suggesting that better communication policies can simultaneously improve task completion and interaction effectiveness. The optimized policies are reproduced in Appendix~\ref{sec:cpe-policies}.

\subsection{Planner--Executor Setting}
\label{sec:planner-executor}

In the Planner--Executor setting (\S\ref{sec:two-settings}), the Planner holds a complete plan $z$ and discloses cooperatively; the Executor receives only a vague summary and must recover missing dimensions through interaction. This removes persona dynamics and task ambiguity, letting us ask: does hybrid communication help even when the information holder is perfectly cooperative and the information need is well-defined?

Table~\ref{tab:planner-executor} reports the results on SWE-bench, TravelGym, and $\tau^2$-bench. The answer is clear: $M_{\text{hybrid}}$ leads in 7 of 10 pairs. Even when the Planner is fully cooperative and the missing information is deterministic, having both channels beats either alone. This reinforces the paper's central claim: communication policy, the meta-decision of \emph{which channel to use}, carries independent value beyond adapting to user personas. When hybrid access underperforms, the information need is narrow enough that a single well-chosen channel suffices; when the need is broader, channel flexibility directly translates to better task outcomes. The hybrid win rate here (7/10) is comparable to the User--Agent setting (8/14, Table~\ref{tab:main-mode-compare}), suggesting that channel selection matters about as much for pure information transmission as it does for navigating human ambiguity.

\section{Conclusion}

In this work, we formalized \emph{Communication Policy}, the prompt-level strategy for choosing between text and UI channels in hybrid agent interaction, and proposed CPE to optimize this decision automatically. Experiments across four environments and two interaction settings reveal two key findings: (i) text and UI communication are complementary rather than interchangeable, with text often supporting more efficient task progression, while structured UI improves response quality and persona compliance; hybrid interaction can combine advantages from both channels and often achieves stronger overall performance; (ii) effective communication strategies can be discovered through prompt-level self-evolution without model retraining. Our results highlight communication behavior as an important and underexplored dimension in LLM agent interaction.
 We hope this work encourages the community to treat communication channel selection as a first-class design concern.

\section*{Limitations}

Our work presents a few limitations that outline directions for future work. First, our experiments use LLM-based user simulators, which enable controlled and reproducible evaluation at scale but may not capture the full variability of real human interaction. Second, our evaluation spans four environments, two interaction settings, and four user personas, but the space of possible communication channel designs remains largely unexplored. Third, our cost annotation protocol assumes users disclose information following a predefined sensitivity schema; real-world disclosure behavior is more complex and context-dependent.

\bibliography{custom}

\appendix

\section{Cost Level Definitions}
\label{sec:cost-defs}

Each benchmark defines a 3--5 level sensitivity hierarchy for the information dimensions in $z$ (\S\ref{sec:partial-info}). The \textit{User Simulator} references these levels to decide what information to disclose at each turn. Table~\ref{tab:cost-all} lists the definitions as written in the \textit{User Simulator} system prompt.

\begin{table*}[t]
\centering
\resizebox{\textwidth}{!}{%
\begin{tabular}{cllll}
\toprule
\textbf{Cost} & \textbf{SWE-bench} & \textbf{TravelGym} & \textbf{$\tau^2$-bench} & \textbf{WebArena} \\
\midrule
1 & Full problem description & Trip framing / scenario text & Known info (account/booking facts) & Task category (site \& action type) \\
2 & Hint information & Dimensions \& specific preferences & Task instructions (goals \& constraints) & Full intent (verbatim objective) \\
3 & Refused / don't know & Refused / don't know & Refused / don't know & Refused / don't know \\
4 & Edit file path & --- & Key identifiers (IDs) & Credentials (login / session / URL) \\
5 & Edit function name & --- & Evaluation criteria (success conditions) & Evaluation criteria (reference answers) \\
\bottomrule
\end{tabular}%
}
\caption{Cost level definitions from the \textit{User Simulator} system prompt. Cost 3 is reserved for refusal across all environments. SWE-bench, $\tau^2$-bench, and WebArena use 5 levels; TravelGym uses 3.}
\label{tab:cost-all}
\end{table*}

\section{\textit{User Simulator} Prompt Structure}
\label{sec:prompt-templates}

The \textit{User Simulator} system prompt contains five sections:

\begin{enumerate}[nosep]
    \item \textbf{Role.} ``You are playing the role of a human user responding to an agent's questions.'' The LLM is instructed to stay in character, sound natural, and never reveal its artificial nature.
    \item \textbf{User Preference \& Reward.} The assigned persona description (e.g., ``The user is an amateur and can only answer very simple and general questions'') and the corresponding reward rule (e.g., ``If the question is very simple, tag [Reward 1]; otherwise, respond I don't know and tag [Reward 0]'').
    \item \textbf{Information \& Cost.} The full task specification $z$ organized into environment-specific sensitivity tiers, each with a defined Cost level. The simulator must tag each response with the highest Cost among the tiers disclosed.
    \item \textbf{Answering Rules.} Default style: concise, casual, natural language; use contractions; single paragraph only; no markdown, bullet points, code blocks, or role explanations; never ask the agent a question.
    \item \textbf{Output Format.} Natural-language reply (single paragraph), then \texttt{[Cost N]} on a new line, then optionally \texttt{[Reward 0/1]} if the reward rule requires it.
\end{enumerate}

The example below is a $\tau^2$-bench \textit{User Simulator} prompt with the amateur persona; the structure is analogous for other benchmarks and personas, with environment-specific cost tiers and persona-specific preference descriptions.
\medskip
\begin{promptbox}[envteal]{$\tau^2$-bench \textit{User Simulator} Prompt (amateur persona)}

\smallskip
\noindent\textbf{System prompt.}

\smallskip
\noindent\emph{Role.}
{\small\texttt{You are playing the role of a human user responding to an agent's questions. Your behavior is dictated by a specific user preference, and you will evaluate the agent's questions based on a reward rule. You will also assign a cost based on the sensitivity of the information you share.}}

\smallskip
\noindent\emph{User Preference \& Reward.}
{\small\ttfamily
Your Preference: The user is an amateur and can only answer very simple and general questions. When a question involves any professional knowledge, the user will respond with ``I don't know.''\\
Reward Rule: If the question is very simple (e.g., related only to personal preference or common sense), tag it as [Reward 1]. Otherwise, if it involves any professional knowledge, respond with ``I don't know.'' and tag it as [Reward 0].}

\smallskip
\noindent\emph{Information \& Cost (airline domain).}
{\small\ttfamily
Known Info (Cost 1): Basic account or booking details the customer knows about themselves.\\
Task Instructions (Cost 2): The customer's detailed goals, constraints, and behavioural directives.\\
Key Identifiers (Cost 4): Specific IDs (user\_id, reservation\_id, etc.).\\
Evaluation Criteria (Cost 5): The exact conditions that must be met for the task to succeed.\\
Cost 3: Information not provided --- refused to answer or said ``I don't know.''\\
~\\
--------------------- TASK INFORMATION ---------------------\\
Known Info: You are Emma Kim. Your user id is emma\_kim\_9957.\\
Task Instructions: You want to cancel reservation EHGLP3. If the agent tells you cancellation is not possible, mention you were told insurance was unnecessary. Do not cancel without a refund.\\
Key Identifiers: None\\
Evaluation Criteria: \{"actions": [], "nl\_assertions": ["Agent should refuse to proceed with the cancellation."], "reward\_basis": ["DB", "COMMUNICATE"]\}}

\smallskip
\noindent\emph{Answering Rules.}
{\small\ttfamily
Adhere to your preference above all else. Default style: keep replies short, casual, and natural. Use contractions (``it's,'' ``I don't''). Always write in a single paragraph. No markdown, bullet points, code blocks, or role explanations. Never ask the agent a question. Escalate to more sensitive information only if the agent is stuck.}

\smallskip
\noindent\emph{Output Format.}
{\small\ttfamily
1. A single-paragraph, human-style, concise reply.\\
2. The cost tag on a new line (e.g., [Cost 3]).\\
3. The reward tag on a new line (e.g., [Reward 1]), only if required by the reward rule.\\
~\\
Again, your preference is: The user is an amateur and can only answer very simple and general questions.\\
Now answer the agent's question. Be very concise. Ensure cost and reward predictions are accurate.}

\end{promptbox}

\section{\textit{Planner Simulator} Prompt Structure}
\label{sec:planner-prompts}

In the Planner--Executor setting (\S\ref{sec:two-settings}), the \textit{User Simulator} is replaced by a \textit{Planner Simulator}: an LLM that holds the full task specification $z$. At the start of each episode, the Planner receives the system prompt (with $z$) and generates a step-by-step execution plan, which is shared with the Executor. On subsequent turns, the Executor sends clarification queries and the Planner answers using the full context. Unlike the \textit{User Simulator}, the Planner has no persona and no reward tagging; it discloses information cooperatively following the cost schema. The prompt consists of four sections:

\begin{enumerate}[nosep]
    \item \textbf{Role.} Frames the LLM as the Planner holding the full ground truth; the Executor reaches the Planner through communication turns. Answers must draw from the ground-truth annex only.
    \item \textbf{Output Format.} A JSON object with \texttt{thought} (private reasoning) and \texttt{response} (what the Executor reads, followed by a Cost tag). No reward tags.
    \item \textbf{Information \& Cost.} Cost tiers (Scenario $\rightarrow$ Cost~1, Dimensions \& Latent Preferences $\rightarrow$ Cost~2, Refusal $\rightarrow$ Cost~3), with concrete TASK INFORMATION appended below: the full scenario, the vague initial\_opening ($\tilde{z}$), which dimensions have latent preferences, and the structured ground-truth constraints.
    \item \textbf{Initial Execution Plan.} A numbered step-by-step plan generated at episode start and shared with the Executor. The Planner must stay consistent with this plan when answering subsequent clarifications.
\end{enumerate}

The example below is a TravelGym \textit{Planner Simulator} prompt; the structure is analogous for other benchmarks, with environment-specific cost tiers and task information.

\medskip
\begin{promptbox}[uiorange]{TravelGym \textit{Planner Simulator} Prompt}

\smallskip
\noindent\textbf{System prompt.}

\smallskip
\noindent\emph{Role.}
{\small\texttt{You are the Planner for this TravelGym episode. You hold the full ground truth in the annex below. The Executor reaches you through action turns (plain text or a UI screenshot). Answer from the annex only; do not invent preferences. Prefer the smallest disclosure that still lets the Executor proceed.}}

\smallskip
\noindent\emph{Output Format.}
{\small\ttfamily
Return exactly one JSON object with two keys:\\
- \textasciigrave thought\textasciigrave: brief private reasoning (may reference strategy).\\
- \textasciigrave response\textasciigrave: what the Executor should read as your reply, then a new line with exactly one of [Cost 1], [Cost 2], or [Cost 3].\\
Do not put tags inside \textasciigrave thought\textasciigrave. Do not add [Reward] lines.}

\smallskip
\noindent\emph{Information \& Cost (TravelGym).}
{\small\ttfamily
Scenario (Cost 1): Trip framing and scenario text --- information already public to the Executor.\\\relax
Dimensions \& Latent Preferences (Cost 2): Structured ground truth beyond the public framing (which dimensions exist, opening line, specific preference lines).\\\relax
Cost 3: Refusal, don't know, or no substantive travel detail.\\
~\\
--------------------- TASK INFORMATION ---------------------\\
Scenario: I am planning a trip to Chicago from November 10th to November 17th. [...] I would love to dine at a restaurant that not only serves authentic American cuisine but also provides convenient parking.\\
Initial opening: I am planning a trip to Chicago from November 10th to November 17th, staying in an apartment. I would like to dine at a restaurant in Chicago on November 12th.\\
Dimensions: restaurant, apartment\\
Latent preferences: [P1] restaurant cuisine -- American $|$ [P2] restaurant parking $|$ [P3] apartment platform -- Airbnb $|$ [P4] apartment rooms -- at least two bathrooms}

\medskip
\noindent\textbf{Initial Execution Plan.}

\smallskip
{\small\ttfamily
1. Confirm travel dates and destination: November 10th--17th, Chicago.\\
2. Search for an Airbnb in Chicago with at least two bathrooms.\\
3. Book the selected Airbnb for the specified dates.\\
4. Research restaurants in Chicago serving authentic American cuisine.\\
5. Identify restaurants that offer convenient parking.\\
6. Select a restaurant meeting both cuisine and parking preferences.\\
7. Make a reservation at the chosen restaurant for November 12th.\\\relax
[...]}

\end{promptbox}

\section{Persona Definitions}
\label{sec:personas}

Experiments use four personas, each assigned to an equal share of episodes across all environments. Each persona is defined by a natural-language preference description injected into the \textit{User Simulator} prompt at the start of an episode, shaping how the user responds to agent questions (or whether they respond at all).

\smallskip
\noindent\textbf{amateur.}
The user is an amateur and can only answer very simple and general questions. When a question involves any professional knowledge, the user responds with ``I don't know.''

\smallskip
\noindent\textbf{do\_selection.}
The user can only answer selection questions. The agent's question must provide options (e.g., A, B, C); the user responds with the chosen letter only. If the question is not a selection question, the user responds with ``I don't know.''

\smallskip
\noindent\textbf{one\_question.}
The user prefers the agent to ask only one question at a time. Multi-part questions receive a refusal.

\smallskip
\noindent\textbf{answer\_more.}
The user prefers the agent to ask more questions. The agent should ask a minimum of 3 questions during the episode.

\smallskip
\noindent The same four preference texts are used across SWE-bench, TravelGym, $\tau^2$-bench, and WebArena.

\subsection{Scoring}

Each persona defines a compliance condition. Let $\text{reward}_0$ be the number of \texttt{[Reward 0]} tags in an episode and $\text{ask\_turn}$ the number of agent questioning turns. The penalty $\text{preference\_reward}$ and compliance flag $\text{preference\_ok}$ are:

\begin{itemize}[nosep]
\item \textbf{amateur}. $\text{preference\_reward} = -0.1 \cdot \text{reward}_0$ (LLM-tagged). Compliant iff no \texttt{[Reward 0]} tags.
\item \textbf{do\_selection}. $\text{preference\_reward} = -0.5 \cdot \text{reward}_0$ (LLM-tagged). Compliant iff all questions provide A/B/C options.
\item \textbf{one\_question}. $\text{preference\_reward} = -0.5 \cdot \text{reward}_0$ (LLM-tagged). Compliant iff each message contains a single question.
\item \textbf{answer\_more}. $\text{preference\_reward} = -\min(\text{ask\_turn} - 3,\; 0)$ (programmatic). Compliant iff $\text{ask\_turn} \geq 3$.
\end{itemize}

\noindent $\text{preference\_ok}=1$ when $\text{preference\_reward}=0$, and 0 otherwise. The personalization score is the mean over successful episodes $\mathcal{E}$:

{\small
\[
\text{Personalization} = \frac{1}{|\mathcal{E}|} \sum_{e \in \mathcal{E}} \text{preference\_ok}(e) \in [0, 1].
\]
}

\section{Environment Details}
\label{sec:env-details}

\begin{table*}[t]
\centering
\footnotesize
\begin{tabular}{p{1.6cm}p{2.8cm}p{2.8cm}p{2.8cm}p{2.8cm}}
\toprule
& \textbf{SWE-bench} & \textbf{TravelGym} & \textbf{$\tau^2$-bench} & \textbf{WebArena} \\
\midrule
Domain & Software repair & Travel planning & Customer service & Web tasks \\[3pt]
$\tilde{z}$ (vague) & ``Fix a login timeout bug'' & ``Plan a weekend trip to Paris'' & ``I need to cancel a booking'' & ``Find and compare products on a shopping site'' \\[3pt]
$z$ (full spec.) & target: \texttt{auth.py}, fn: \texttt{validate\_session}, handle null token & budget: \texteuro300, hotel: 3-star Louvre, cuisine: Italian, direct flight & reservation: 8HG87, refund: visa 4456, tier: gold & search: ``Red Coffee Mug,'' add to cart, verify price \$12.99 \\[3pt]
Hidden info & Hints, file paths, function names & Preferences, constraints, GT items & Account info, task instructions, IDs, eval criteria & Intent details, values, credentials \\[3pt]
Cost levels & 1--5 & 1--3 & 1--5 & 1--5 \\[3pt]
Success signal & Patch similarity & Terminal reward & DB state match & URL/element match \\
\bottomrule
\end{tabular}
\caption{Evaluation environments with example vague specifications and task specifications. The agent sees only $\tilde{z}$; the full $z$ is held by the \textit{User Simulator} or \textit{Planner Simulator}.}
\label{tab:envs}
\end{table*}

In all benchmarks, for $M_{\text{ui}}$ and $M_{\text{hybrid}}$ modes, \texttt{generate\_ui} HTML is rendered via Playwright as a screenshot and sent to the multimodal user LLM. \texttt{ask\_question} sends plain text.

\smallskip
\noindent Table~\ref{tab:envs} gives an example vague–full specification pair for each benchmark. The per-benchmark vaguification rules are as follows.

\subsection{SWE-bench}
We use SWE-bench~\citep{jimenez2024swe}, a repository-level software repair benchmark. We evaluate on 56~tasks from SWE-bench Verified, each paired with 4~personas (amateur, answer\_more, do\_selection, one\_question), yielding 224~episodes per experimental configuration. Max~turns is 200.

The agent debugs and edits code in a Linux sandbox with environment tools \texttt{execute\_bash} (read-only) and \texttt{str\_replace\_editor}. Communication tools \texttt{ask\_question} and \texttt{generate\_ui} are interleaved with code operations. Success is measured by the F1 similarity between the agent's final diff and the oracle patch.

\smallskip
\noindent\textbf{Vaguification.} $z$ contains the full issue description, developer hints, the target file path(s), and the target function/class name(s). $\tilde{z}$ is a \texttt{summarized\_issue}: a one-paragraph abbreviation that drops file paths, function names, and hint details.

\subsection{TravelGym}
We use TravelGym~\citep{qian2025userbench}, a multi-turn travel planning benchmark. We evaluate on 50~tasks, each paired with 4~personas, yielding 200~episodes per configuration. Max~turns is 20.

The agent uses a single \texttt{interact\_with\_env} tool with sub-commands \texttt{search}, \texttt{answer}, \texttt{finish}, and \texttt{action}. Communication is multiplexed through \texttt{action}: \texttt{ask\_question} corresponds to \texttt{content\_kind=text}, and \texttt{generate\_ui} to \texttt{content\_kind=html}. Success is measured by the environment's native terminal reward (max~2.4), combining an LLM judge evaluation of the final itinerary with rule-based answer matching against ground truth.

\smallskip
\noindent\textbf{Vaguification.} $z$ contains the full trip scenario, the list of travel dimensions (flight, hotel, restaurant), per-dimension latent preference constraints (e.g., cabin class, star rating, cuisine type), and the ground-truth answer IDs. $\tilde{z}$ is the \texttt{initial\_description}: a 1--2 sentence user opening (e.g., ``Plan a weekend trip to Paris'') that drops all dimension-level preferences and constraints.

\subsection{$\tau^2$-bench}
We use $\tau^2$-bench~\citep{barres2025tau} in the airline domain only. We evaluate on 50~tasks, each paired with 4~personas, yielding 200~episodes per configuration. Max~turns is 100.

The agent uses domain-specific function calls (\texttt{book\_reservation}, \texttt{search\_direct\_flight}, \texttt{get\_user\_details}, etc.) to manipulate a simulated airline database. Communication tools \texttt{ask\_question} and \texttt{generate\_ui} are appended to the domain tool list; \texttt{ask\_question} calls are intercepted and routed to the \textit{User Simulator}. Success is binary: the final database state must exactly match the expected state from a reference execution.

\smallskip
\noindent\textbf{Vaguification.} $z$ contains known account/booking facts (\texttt{known\_info}), customer task instructions (\texttt{task\_instructions}), key identifiers (\texttt{user\_id}, \texttt{reservation\_id}), and evaluation criteria. $\tilde{z}$ is the \texttt{reason\_for\_call}: the customer's opening message (1--3 sentences, e.g., ``I need to cancel a booking''), which omits all account details, identifiers, and success conditions.

\subsection{WebArena}
We use WebArena~\citep{zhou2024webarena} across 3~sites: shopping, Reddit, and GitLab (map and wiki are excluded). We evaluate on 50~tasks, each paired with 4~personas, yielding 200~episodes per configuration. Max~turns is 50.

The agent navigates live web environments via browser primitives (\texttt{goto}, \texttt{click}, \texttt{fill}, \texttt{scroll}, \texttt{select\_option}). Communication tools \texttt{ask\_question} and \texttt{generate\_ui} are added alongside browser actions. Each LLM call receives a single system message plus a flat task block (current URL, previous action, page accessibility tree). Success is evaluated by URL match, page element presence, or expected string content on the final page.

\smallskip
\noindent\textbf{Vaguification.} $z$ contains the full intent (verbatim task description), target sites, credentials (\texttt{username}, \texttt{password}, \texttt{start\_url}), and evaluation criteria (reference answers, URL checks). $\tilde{z}$ is a one-sentence summary that infers the task category (e.g., purchase, search, post) from the intent via keyword matching and maps site names to short descriptions (e.g., ``I need to purchase something on an e-commerce shopping website''). All item names, prices, dates, and credential details are dropped.

\section{Communication Policy Prompts}
\label{sec:comm-policy-prompts}

The communication policy $\pi_{\text{comm}}$ is the prompt text rendered to the agent at the start of each episode. It consists of three components: (1)~the system prompt, specifying the agent's role, available tools, and channel-selection rules; (2)~few-shot examples demonstrating when to use each channel; and (3)~environment-specific guidance (e.g., task framing, cost schemas). Below we provide the policy for each benchmark as used in our experiments.

\subsection{SWE-bench}

\medskip
\begin{promptbox}[agentblue]{SWE-bench Communication Policy}

\smallskip
\noindent\textbf{System prompt.}

\smallskip
\noindent\emph{Role.} {\small\texttt{You are OpenHands agent, a helpful AI assistant that can interact with a computer to solve tasks.}}

\smallskip
\noindent\emph{Environment tools.}
\begin{itemize}[nosep,leftmargin=2em]
  \item {\small\texttt{execute\_bash}}
  \item {\small\texttt{finish}}
  \item {\small\texttt{str\_replace\_editor}}
\end{itemize}

\smallskip
\noindent\emph{Communication tools.}
\begin{itemize}[nosep,leftmargin=2em]
  \item {\small\texttt{ask\_question}}
  \item {\small\texttt{generate\_ui}}
\end{itemize}

\smallskip
\noindent\emph{Channel-selection rules.}
\begin{itemize}[nosep,leftmargin=2em]
  \item {\small\texttt{Do not ask the user in free-form assistant text; use either ask\_question (plain text) or generate\_ui (visual HTML).}}
  \item {\small\texttt{Use ask\_question when a short text question is enough.}}
  \item {\small\texttt{Use generate\_ui when a rendered page helps (forms, options, layout); html\_code must be a complete valid HTML5 document.}}
  \item {\small\texttt{Adapt to the user's preference in the task description; be creative with HTML.}}
\end{itemize}

\medskip
\noindent\textbf{Examples.}

\smallskip
\noindent\emph{Example~A — text (ask\_question).}
{\small\ttfamily
~\\
USER: The app sometimes shows a random error page.
Can you figure out what's wrong?\\
~\\
ASSISTANT: I'll list the top-level layout, then ask one focused question if details are missing.\\
<function=execute\_bash>
<parameter=command>ls -1 /testbed</parameter>
</function>\\
~\\
USER: api~~frontend~~scripts~~tests\\
~\\
ASSISTANT:\\
<function=ask\_question>
<parameter=query>What is the exact error message or
stack trace when the error page appears?</parameter>
</function>
}

\smallskip
\noindent\emph{Example~B — UI (generate\_ui).}
{\small\ttfamily
~\\
ASSISTANT (needs more structured input):\\
<function=generate\_ui>\\
<parameter=progress\_summary>\\I've mapped the repo.
I need one concrete detail to reproduce the issue.</parameter>
<parameter=html\_code><!doctype html>
<html><head><meta charset="utf-8"/><title>Reproduce the issue</title>
[...] <label>Timing</label>
<select><option>On first load</option> [...] </select>
<label>What do you see?</label>
<textarea rows="3" placeholder="Error text or steps"></textarea>
[...] </html></parameter>
<parameter=questions>[{"question":
"When does the error occur and what appears on screen?"}]
</parameter>
</function>
}

\medskip
\noindent\textbf{Appendixes.}

\smallskip
{\small\ttfamily
NEW TASK DESCRIPTION: The app sometimes shows
a random error page. Can you figure out what's wrong?
~\\
The user's preference for the agent is: The user is an
amateur and can only answer very simple and general
questions. When a question involves any professional
knowledge, the user will respond with "I don't know."
}

\end{promptbox}

\subsection{TravelGym}

\medskip
\begin{promptbox}[uiorange]{TravelGym Communication Policy}

\smallskip
\noindent\textbf{System prompt.}

\smallskip
\noindent\emph{Role.} {\small\texttt{You are an agent that actively interacts with a specific environment.}}

\smallskip
\noindent\emph{Environment tools.}
\begin{itemize}[nosep,leftmargin=2em]
  \item {\small\texttt{search}}
  \item {\small\texttt{answer}}
  \item {\small\texttt{finish}}
\end{itemize}

\smallskip
\noindent\emph{Communication tools.}
\begin{itemize}[nosep,leftmargin=2em]
  \item {\small\texttt{ask\_question}}
  \item {\small\texttt{generate\_ui}}
\end{itemize}

\smallskip
\noindent\emph{Channel-selection rules.}
\begin{itemize}[nosep,leftmargin=2em]
  \item {\small\texttt{Do not ask in free-form assistant text; use interact\_with\_env (text) or interact\_with\_env (html).}}
  \item {\small\texttt{Use content\_kind=text when a short text question is enough.}}
  \item {\small\texttt{Use content\_kind=html when a rendered page helps (forms, options, layout).}}
  \item {\small\texttt{Adapt to the user's preference in the task description; be creative with HTML.}}
\end{itemize}

\smallskip
\noindent\emph{Behavior rules.}
{\small\ttfamily
The user's travel needs may be vague or underspecified.
When you need clarification, use interact\_with\_env
with choice=action and a content\_kind prefix to select
the channel. Each question should address a specific
missing preference dimension (budget, cuisine, star
rating, etc.).
}

\medskip
\noindent\textbf{Examples.}

\smallskip
\noindent\emph{Example~A — text (ask\_question).}
{\small\ttfamily
~\\
USER: I am planning a trip to Bangkok, staying in an
apartment from November 10th to November 17th, and
visiting a restaurant on November 11th. Also my budget
is limited...\\
~\\
ASSISTANT (after searching for apartments):\\
<function=interact\_with\_env>\\
<parameter=choice>action</parameter>\\
<parameter=content\_kind>text</parameter>\\
<parameter=content>Do you have a preference for a
specific area in Bangkok, or is any area fine as
long as it's within budget?</parameter>
</function>
}

\smallskip
\noindent\emph{Example~B — UI (generate\_ui).}
{\small\ttfamily
~\\
ASSISTANT (needs to narrow down restaurant options):\\
<function=interact\_with\_env>\\
<parameter=choice>action</parameter>\\
<parameter=content\_kind>html</parameter>\\
<parameter=content><!doctype html>
<html><head><meta charset="utf-8"/><title>Restaurant Preferences</title>
[...] <label>Cuisine type</label>
<select><option>Thai</option> [...] </select>
<label>Budget per person (THB)</label>
<select><option>Under 300</option> [...] </select>
[...] </html></parameter>
</function>
}

\medskip
\noindent\textbf{Appendixes.}

\smallskip
{\small\ttfamily
I am planning a business trip to Austin, Texas, from
April 15th to April 20th, staying in an apartment. I
also need a restaurant reservation in Austin on April
18th. \\
~\\
The user's preference for the agent is: The user is an
amateur and can only answer very simple and general
questions. When a question involves any professional
knowledge, the user will respond with "I don't know."
}

\end{promptbox}

\subsection{$\tau^2$-bench}

\medskip
\begin{promptbox}[envteal]{$\tau^2$-bench Communication Policy}

\smallskip
\noindent\textbf{System prompt.}

\smallskip
\noindent\emph{Role.} {\small\texttt{You are a professional customer service agent for an airline company. Your job is to resolve the customer's issue efficiently and accurately, following company policy.}}

\smallskip
\noindent\emph{Environment tools.}
\begin{itemize}[nosep,leftmargin=2em]
  \item {\small\texttt{book\_reservation}}
  \item {\small\texttt{cancel\_reservation}}
  \item {\small\texttt{search\_direct\_flight}}
  \item {\small\texttt{search\_onestop\_flight}}
  \item {\small\texttt{get\_user\_details}}
  \item {\small\texttt{get\_reservation\_details}}
  \item {\small\texttt{get\_flight\_status}}
  \item {\small\texttt{list\_all\_airports}}
  \item {\small\texttt{send\_certificate}}
  \item {\small\texttt{update\_reservation\_flights}}
  \item {\small\texttt{update\_reservation\_passengers}}
  \item {\small\texttt{update\_reservation\_baggages}}
  \item {\small\texttt{calculate}}
  \item {\small\texttt{transfer\_to\_human\_agents}}
\end{itemize}

\smallskip
\noindent\emph{Communication tools.}
\begin{itemize}[nosep,leftmargin=2em]
  \item {\small\texttt{ask\_question}}
  \item {\small\texttt{generate\_ui}}
\end{itemize}

\smallskip
\noindent\emph{Channel-selection rules.}
\begin{itemize}[nosep,leftmargin=2em]
  \item {\small\texttt{Do not ask the user in free-form assistant text; use either ask\_question (plain text) or generate\_ui (visual HTML).}}
  \item {\small\texttt{Use ask\_question when a short text question is enough.}}
  \item {\small\texttt{Use generate\_ui when a rendered page helps (forms, options, layout); html\_code must be a complete valid HTML5 document.}}
  \item {\small\texttt{Adapt to the user's preference in the task description; be creative with HTML.}}
\end{itemize}

\medskip
\noindent\textbf{Examples.}

\smallskip
\noindent\emph{Example~A — text (ask\_question).}
{\small\ttfamily
CUSTOMER: I need to cancel a booking.\\
~\\
ASSISTANT:\\
<function=ask\_question>\\
<parameter=query>Could you provide the reservation ID
or the full name on the booking?</parameter>
</function>
}

\smallskip
\noindent\emph{Example~B — UI (generate\_ui).}
{\small\ttfamily
ASSISTANT (needs to confirm flight change details):\\
<function=generate\_ui>\\
<parameter=progress\_summary>I found your reservation.
To change your flight, I need a few details.</parameter>\\
<parameter=html\_code><!doctype html>
<html><head><meta charset="utf-8"/><title>Change Flight</title>
[...] <label>New departure date</label>
<input type="date" name="new\_date"/>
<label>Preferred time</label>
<select><option>Morning</option> [...] </select>
[...] </html></parameter>
<parameter=questions>[{"question":
"What date and time would you prefer
for your new flight?"}]</parameter>
</function>
}

\medskip
\noindent\textbf{Appendixes.}

\smallskip
{\small\ttfamily
NEW TASK DESCRIPTION: A customer has contacted
customer service. Here is their initial message:\\
--- BEGIN CUSTOMER MESSAGE ---\\
I recently spoke on the phone with a customer support
representative that told me that a service agent will
be able to help me cancel my reservation.\\
--- END CUSTOMER MESSAGE ---\\
Resolve this issue.\\
~\\
The customer's preference: The user can only answer
selection questions. The agent's question must provide
options such as A, B, C, and the user will respond
with their choice (e.g., A).
}

\end{promptbox}

\subsection{WebArena}

\medskip
\begin{promptbox}[hybridpurple]{WebArena Communication Policy}

\smallskip
\noindent\textbf{System prompt.}

\smallskip
\noindent\emph{Role.} {\small\texttt{You are a web automation agent. You control a real web browser to complete tasks on behalf of the user.}}

\smallskip
\noindent\emph{Environment tools.}
\begin{itemize}[nosep,leftmargin=2em]
  \item {\small\texttt{goto}}
  \item {\small\texttt{click}}
  \item {\small\texttt{fill}}
  \item {\small\texttt{press}}
  \item {\small\texttt{scroll}}
  \item {\small\texttt{select\_option}}
\end{itemize}

\smallskip
\noindent\emph{Communication tools.}
\begin{itemize}[nosep,leftmargin=2em]
  \item {\small\texttt{ask\_question}}
  \item {\small\texttt{generate\_ui}}
\end{itemize}

\smallskip
\noindent\emph{Channel-selection rules.}
\begin{itemize}[nosep,leftmargin=2em]
  \item {\small\texttt{Do not ask in free-form assistant text; use ask\_question (plain text) or generate\_ui (visual HTML).}}
  \item {\small\texttt{Use ask\_question when a short text question is enough.}}
  \item {\small\texttt{Use generate\_ui when a rendered page helps (forms, options, layout); html\_code must be a complete valid HTML5 document.}}
    \item {\small\texttt{Adapt to the user's preference in the task description; be creative with HTML.}}
\end{itemize}

\medskip
\noindent\textbf{Examples.}

\smallskip
\noindent\emph{Example~A — text (ask\_question).}
{\small\ttfamily
\\
TASK: I need to find some information on an\\
e-commerce admin panel.\\
~\\
ASSISTANT (after navigating to the admin page):
<function=ask\_question>\\
<parameter=query>What specific information are you
looking for -- sales reports, customer data,
or product catalog details?</parameter>
</function>
}

\smallskip
\noindent\emph{Example~B — UI (generate\_ui).}
{\small\ttfamily
\\
ASSISTANT (needs search criteria):\\
<function=generate\_ui>\\
<parameter=progress\_summary>I'm on the admin panel.
I need to know what to search for.</parameter>]]
<parameter=html\_code><!doctype html>
<html><head><meta charset="utf-8"/><title>Search Criteria</title>
[...] <label>What are you looking for?</label>
<select><option>Sales report</option> [...] </select>
<label>Any specific date range?</label>
<input type="text" placeholder="e.g., Last 7 days"/>
[...] </html></parameter>\\
<parameter=questions>[{"question":\\
"What data do you need and for what time period?"}]
</parameter>
</function>
}

\medskip
\noindent\textbf{Appendixes.}

\smallskip
{\small\ttfamily
Task: I need to purchase something on an e-commerce
shopping website.\\
~\\
The user's preference for the agent is: The user can
only answer selection questions. The agent's question
must provide options such as A, B, C, and the user
will respond with their choice (e.g., A).
}

\end{promptbox}

\section{Evolution Reflect Prompts}
\label{sec:reflect-prompts}

In each Evolve step (Section~\ref{sec:cpe}), an LLM is prompted to analyze rollout results and propose policy edits as a JSON patch. The prompt consists of a system message followed by seven numbered sections: \S1 optimization goals, \S2 modification requirements, \S3 output JSON schema, \S4 per-task scores, \S5 current policy snapshots, \S6 reference trajectories, and \S7 a final formatting instruction. Sections~4--6 are populated dynamically from the current round's rollout data. Below we show the full template for each benchmark; static sections are reproduced as written, while dynamic sections use \texttt{[...]} to indicate the content they carry.

\subsection{SWE-bench}

\smallskip
\begin{promptbox}[agentblue]{SWE-bench Reflect Prompt}

\smallskip
\noindent\textbf{System prompt.}
{\small\ttfamily
You are a senior prompt engineer. The downstream agent may use ask\_question (plain text) or generate\_ui (HTML) to obtain user information. You must respond with one JSON object only (no markdown fences), matching the schema in the user message.\\
The user message supplies schemas, scores, and the prompts to revise. Improve guidance so the agent chooses ask\_question vs generate\_ui appropriately.}

\smallskip
\noindent\emph{\S1 Optimization goals.}
{\small\ttfamily
Your objective is better end-task performance by editing prompts so the agent chooses appropriately between ask\_question (quick text clarification) and generate\_ui (structured HTML) when it needs user input.\\
- System / example prompts: teach when to use plain text vs a rendered page, how much to inspect first, and how to keep clarifications focused. Do not contradict the frozen tool catalog / XML template block.\\
- generate\_ui: use when multi-field, branching, or layout clearly helps the user answer.\\
- ask\_question: use when a single short answer is enough.\\
- SWE: reduce wasted edits before key facts; align with repo exploration norms.}

\smallskip
\noindent\emph{\S2 Modification requirements.}
{\small\ttfamily
- Improve prompts using the score table (this round's eval slice) and the reference trajectories in section 6.\\
- Prefer \textasciigrave system\_delta\textasciigrave\ / \textasciigrave example\_delta\textasciigrave\ when small additions suffice; you may still output full \textasciigrave hybrid\_system\textasciigrave\ / \textasciigrave hybrid\_example\textasciigrave\ if needed.\\
- \textasciigrave hybrid\_system\textasciigrave\ / \textasciigrave system\_delta\textasciigrave: the tool block from \textasciigrave You have access to the following functions:\textasciigrave\ through the first format-example \textasciigrave </function>\textasciigrave\ before \textasciigrave <IMPORTANT>Reminder:\textasciigrave\ must remain byte-identical to the current snapshot.\\
- Do not include: gold answers, hidden test labels, \textasciigrave preference\_ok\textasciigrave\ shortcuts, or dataset leakage.\\
- Output valid JSON only (one top-level object), UTF-8 strings.}

\smallskip
\noindent\emph{\S3 Output JSON schema.}
{\small\texttt{\{"utils": \{"hybrid\_system": "<optional full replacement string>", "hybrid\_example": "<optional full replacement string>", "system\_delta": "<optional suffix appended to current hybrid system>", "example\_delta": "<optional suffix appended to current hybrid example>"\}\}}}

\smallskip
\noindent\emph{\S4 Per-task scores.}
{\small\texttt{[For each task $i \in \mathcal{B}$, hybrid-only scores on productivity.]}}

\smallskip
\noindent\emph{\S5 Current policy snapshots.}
{\small\texttt{[Full text of the current \textasciigrave hybrid\_system\textasciigrave\ and \textasciigrave hybrid\_example\textasciigrave\ prompts, incorporating all prior accepted edits.]}}

\smallskip
\noindent\emph{\S6 Reference trajectories.}
{\small\texttt{[For each task, hybrid rollout excerpts showing every \textasciigrave ask\_question\textasciigrave\ and \textasciigrave generate\_ui\textasciigrave\ call with the user-simulator response, preserving turn order.]}}

\smallskip
\noindent\emph{\S7 Final instruction.}
{\small\texttt{Emit one JSON object as in section 3 (JSON schema). No markdown outside that JSON.}}

\end{promptbox}

\subsection{TravelGym}

\medskip
\begin{promptbox}[uiorange]{TravelGym Reflect Prompt}

\smallskip
\noindent\textbf{System prompt.}
{\small\ttfamily
You are a senior prompt engineer. The downstream agent may use ask\_question (plain text) or generate\_ui (HTML) to obtain user information. You must respond with one JSON object only (no markdown fences), matching the schema in the user message.\\
Target environment: TravelGym / preference-driven travel. Agent prompts come from the dataset parquet (\textasciigrave interact\_with\_env\textasciigrave, etc.); you improve behavior by emitting \textasciigrave travel.system\_suffix\textasciigrave\ only (see schema).}

\smallskip
\noindent\emph{\S1 Optimization goals.}
{\small\ttfamily
Your objective is better end-task performance by editing \textasciigrave travel.system\_suffix\textasciigrave\ only (plain text appended each round to the Travel agent's first system message).\\
- Teach when to use ask\_question vs generate\_ui, and how to use \textasciigrave interact\_with\_env\textasciigrave\ (\textasciigrave choice=action\textasciigrave, \textasciigrave content\_kind=text|html\textasciigrave) appropriately.\\
- Suffixes accumulate across rounds; prefer small, non-contradictory additions.\\
- Travel: respect stated amateur/pro tone; avoid over-proactive suggestions.}

\smallskip
\noindent\emph{\S2 Modification requirements.}
{\small\ttfamily
- Improve behavior using the score table (this round's eval slice) and the reference trajectories in section 6.\\
- Output \textasciigrave travel.system\_suffix\textasciigrave\ only: appended to the first system message. Omit \textasciigrave travel.user\_suffix\textasciigrave\ unless you have a rare cross-task user-channel policy (discouraged).\\
- Prefer short suffixes (tone, when to use \textasciigrave ask\_question\textasciigrave\ vs \textasciigrave generate\_ui\textasciigrave, \textasciigrave interact\_with\_env\textasciigrave\ with \textasciigrave choice=action\textasciigrave\ and \textasciigrave content\_kind=text|html\textasciigrave).\\
- Do not include: gold answers, hidden test labels, or dataset leakage.\\
- Output valid JSON only (one top-level object), UTF-8 strings.}

\smallskip
\noindent\emph{\S3 Output JSON schema.}
{\small\texttt{\{"travel": \{"system\_suffix": "<string appended to first agent system>"\}\}}}

\smallskip
\noindent\emph{\S4 Per-task scores.}
{\small\texttt{[For each task $i \in \mathcal{B}$, hybrid-only scores on productivity.]}}

\smallskip
\noindent\emph{\S5 Current policy snapshots.}
{\small\texttt{[The native Travel agent system prompt as rebuilt by the eval pipeline.}}

\smallskip
\noindent\emph{\S6 Reference trajectories.}
{\small\texttt{[For each task, hybrid rollout excerpts showing every \textasciigrave interact\_with\_env\textasciigrave\ call (tagged with \textasciigrave content\_kind=text|html\textasciigrave) and the user-simulator response.]}}

\smallskip
\noindent\emph{\S7 Final instruction.}
{\small\texttt{Emit one JSON object as in section 3 (JSON schema). No markdown outside that JSON.}}

\end{promptbox}

\subsection{$\tau^2$-bench}

\medskip
\begin{promptbox}[envteal]{$\tau^2$-bench Reflect Prompt}

\smallskip
\noindent\textbf{System prompt.}
{\small\ttfamily
You are a senior prompt engineer. The downstream agent may use ask\_question (plain text) or generate\_ui (HTML) to obtain user information. You must respond with one JSON object only (no markdown fences), matching the schema in the user message.\\
Target environment: Tau2 / airline customer-service assistant. Agent system is rebuilt in-code per domain; you improve behavior by emitting \textasciigrave tau2.system\_suffix\textasciigrave\ only (see schema). Nudge ask\_question vs generate\_ui and keep clarifications focused and domain-appropriate.}

\smallskip
\noindent\emph{\S1 Optimization goals.}
{\small\ttfamily
Your objective is better end-task performance by editing \textasciigrave tau2.system\_suffix\textasciigrave\ only (plain text appended each round to the $\tau^2$ agent's rebuilt system in hybrid mode).\\
- Teach when to use ask\_question (short text) vs generate\_ui (HTML), and how to keep clarifications focused.\\
- Suffixes accumulate across rounds; prefer small, non-contradictory additions.\\
- Do not attempt to replace the in-code $\tau^2$ tool catalog or \textasciigrave <function=\textasciigrave\ XML templates via JSON; use suffixes for policy and tone only.\\
- Tau2: keep clarifications domain-neutral unless the task domain is explicit.}

\smallskip
\noindent\emph{\S2 Modification requirements.}
{\small\ttfamily
- Improve behavior using the score table (this round's eval slice) and the reference trajectories in section 6.\\
- Output \textasciigrave tau2.system\_suffix\textasciigrave\ only: a plain string appended to the $\tau^2$ agent's rebuilt system for the next eval. Omit \textasciigrave tau2.user\_suffix\textasciigrave\ unless you have a rare cross-task user-channel policy (discouraged).\\
- Prefer short suffixes; do not try to replace the in-code tool or XML templates---suffixes are for policy and tone only.\\
- Do not include: gold answers, hidden test labels, \textasciigrave preference\_ok\textasciigrave\ shortcuts, or dataset leakage.\\
- Output valid JSON only (one top-level object), UTF-8 strings.}

\smallskip
\noindent\emph{\S3 Output JSON schema.}
{\small\texttt{\{"tau2": \{"system\_suffix": "<string appended to agent system; prefer non-empty when you have a concrete improvement>"\}\}}}

\smallskip
\noindent\emph{\S4 Per-task scores.}
{\small\texttt{[For each task $i \in \mathcal{B}$, hybrid-only scores on productivity.]}}

\smallskip
\noindent\emph{\S5 Current policy snapshots.}
{\small\texttt{[The $\tau^2$ agent system prompt as rebuilt by the eval pipeline.]}}

\smallskip
\noindent\emph{\S6 Reference trajectories.}
{\small\texttt{[For each task, hybrid rollout excerpts showing every \textasciigrave ask\_question\textasciigrave\ and \textasciigrave generate\_ui\textasciigrave\ call with the user-simulator response.]}}

\smallskip
\noindent\emph{\S7 Final instruction.}
{\small\texttt{Emit one JSON object as in section 3 (JSON schema). No markdown outside that JSON.}}

\end{promptbox}

\subsection{WebArena}

\medskip
\begin{promptbox}[hybridpurple]{WebArena Reflect Prompt}

\smallskip
\noindent\textbf{System prompt.}
{\small\ttfamily
You are a senior prompt engineer. The downstream agent may use ask\_question (plain text) or generate\_ui (HTML) to obtain user information. You must respond with one JSON object only (no markdown fences), matching the schema in the user message.\\
Target environment: WebArena / browser-style tasks. For\_genui\textasciigrave; for \textasciigrave env\_family=webarena\textasciigrave\ you emit \textasciigrave webarena.system\_suffix\textasciigrave\ only (see schema).}

\smallskip
\noindent\emph{\S1 Optimization goals.}
{\small\ttfamily
Your objective is better end-task performance by editing \textasciigrave webarena.system\_suffix\textasciigrave\ only (plain text appended each round to the WebArena agent's rebuilt system in GenUI hybrid mode).\\
- Teach when to use ask\_question vs generate\_ui after inspecting the page, and how to keep clarifications task-grounded.\\
- Suffixes accumulate across rounds; prefer small, non-contradictory additions.\\
- Do not try to replace the full browser tool XML block or inject OpenHands repo tools via JSON.\\
- WebArena: prefer choices grounded in visible page/task state; never leak evaluation labels.}

\smallskip
\noindent\emph{\S2 Modification requirements.}
{\small\ttfamily
- Improve behavior using the score table (this round's eval slice) and the reference trajectories in section 6.\\
- Output \textasciigrave webarena.system\_suffix\textasciigrave\ only: appended to the rebuilt system. Omit \textasciigrave webarena.user\_suffix\textasciigrave\ unless you have a rare cross-task user-channel policy (discouraged).\\
- Prefer short suffixes (policy, tone, ask vs UI).\\
- Do not include: gold answers, hidden test labels, or dataset leakage.\\
- Output valid JSON only (one top-level object), UTF-8 strings.}

\smallskip
\noindent\emph{\S3 Output JSON schema.}
{\small\texttt{\{"webarena": \{"system\_suffix": "<string appended to rebuilt agent system>"\}\}}}

\smallskip
\noindent\emph{\S4 Per-task scores.}
{\small\texttt{[For each task $i \in \mathcal{B}$, hybrid-only scores on productivity.]}}

\smallskip
\noindent\emph{\S5 Current policy snapshots.}
{\small\texttt{[The WebArena agent system prompt as rebuilt by the eval pipeline.]}}

\smallskip
\noindent\emph{\S6 Reference trajectories.}
{\small\texttt{[For each task, hybrid rollout excerpts showing every \textasciigrave ask\_question\textasciigrave\ and \textasciigrave generate\_ui\textasciigrave\ call with the user-simulator response.]}}

\smallskip
\noindent\emph{\S7 Final instruction.}
{\small\texttt{Emit one JSON object as in section 3 (JSON schema). No markdown outside that JSON.}}

\end{promptbox}

\section{Persona-Level Case Studies}
\label{sec:persona-case}

We present persona-level analyses that illustrate how communication mode effects are moderated by user persona. All values are productivity / proactivity / personalization; best productivity per row in \textcolor{agentblue}{bold blue}.

\subsection{Persona Determines the Best Mode}

\begin{table}[htbp]
\centering
\resizebox{\columnwidth}{!}{%
\begin{tabular}{lcccc}
\toprule
\textbf{Persona} & \textbf{$M_{\text{text}}$} & \textbf{$M_{\text{ui}}$} & \textbf{$M_{\text{hybrid}}$} \\
\midrule
amateur      & \textcolor{agentblue}{\textbf{.260}}/.220/.640 & .160/.140/.780 & .140/.000/.480 \\
answer\_more & .240/.180/.960 & \textcolor{agentblue}{\textbf{.320}}/.020/.920 & \textcolor{agentblue}{\textbf{.320}}/.000/1.00 \\
do\_selection & \textcolor{agentblue}{\textbf{.380}}/.240/.460 & .320/.040/.740 & \textcolor{agentblue}{\textbf{.380}}/.020/.640 \\
one\_question & .200/.280/.540 & \textcolor{agentblue}{\textbf{.300}}/.280/.960 & .240/.040/.320 \\
\bottomrule
\end{tabular}%
}
\caption{$\tau^2$-bench DeepSeek-V3.2 + GPT-4o: productivity / proactivity / personalization by persona. The best mode varies with persona---text for amateur, UI/hybrid for answer\_more, text/hybrid for do\_selection, UI for one\_question.}
\label{tab:case-tau2-dsv3}
\end{table}

Table~\ref{tab:case-tau2-dsv3} breaks down $\tau^2$-bench DeepSeek-V3.2 + GPT-4o by persona. The optimal communication mode differs for every persona: amateur users achieve the highest productivity with $M_{\text{text}}$ (.260), answer\_more with either $M_{\text{ui}}$ or $M_{\text{hybrid}}$ (.320), do\_selection with $M_{\text{text}}$ or $M_{\text{hybrid}}$ (.380), and one\_question with $M_{\text{ui}}$ (.300). No single channel dominates across personas, directly validating the paper's core claim that optimal communication policy is conditional on \emph{who} the user is.

\subsection{When Naive Hybrid Access Backfires}

\begin{table}[htbp]
\centering
\resizebox{\columnwidth}{!}{%
\begin{tabular}{lcccc}
\toprule
\textbf{Persona} & \textbf{$M_{\text{text}}$} & \textbf{$M_{\text{ui}}$} & \textbf{$M_{\text{hybrid}}$} \\
\midrule
amateur      & .948/.380/.820  & \textcolor{agentblue}{\textbf{1.012}}/.400/.900  & .780/.100/.525 \\
answer\_more & \textcolor{agentblue}{\textbf{1.328}}/.340/.880 & 1.040/.160/1.00  & 1.116/.260/.740 \\
do\_selection & .788/.460/.760  & \textcolor{agentblue}{\textbf{.956}}/.220/.580  & .796/.300/.878 \\
one\_question & \textcolor{agentblue}{\textbf{.988}}/.200/.940 & .972/.360/.980  & .932/.120/.925 \\
\bottomrule
\end{tabular}%
}
\caption{TravelGym GPT-5-mini + Qwen3-VL-32B: productivity / proactivity / personalization by persona. $M_{\text{hybrid}}$ trails at least one single-channel baseline for amateur, answer\_more, and one\_question.}
\label{tab:case-travelgym-gpt5mini}
\end{table}

Table~\ref{tab:case-travelgym-gpt5mini} breaks down TravelGym GPT-5-mini + Qwen3-VL-32B by persona. $M_{\text{hybrid}}$ underperforms at least one single-channel baseline in three of four personas, with the amateur persona showing the steepest drop ($M_{\text{ui}}$ 1.012 $\rightarrow$ $M_{\text{hybrid}}$ .780, $-23\%$). When the agent lacks the judgment to choose channels wisely, having both channels is worse than having one, directly motivating the need for CPE.

\subsection{Cross-Benchmark Persona--Mode Alignment}

The two analyses above illustrate persona-mode interactions in specific settings. Here we compare across all 14 agent--user experiment configurations to identify systematic persona--benchmark--mode alignment patterns.

\smallskip
\noindent\textbf{Case 1: Amateur users. $M_{\text{text}}$ is the safest choice.}

\begin{table}[htbp]
\centering
\resizebox{\columnwidth}{!}{%
\begin{tabular}{lccc}
\toprule
\textbf{Benchmark} & \textbf{$M_{\text{text}}$} & \textbf{$M_{\text{ui}}$} & \textbf{$M_{\text{hybrid}}$} \\
\midrule
$\tau^2$-bench & \textcolor{agentblue}{\textbf{.320}}/.060/.385 & .255/\textcolor{uiorange}{\textbf{.065}}/\textcolor{envteal}{\textbf{.545}} & .250/.035/.394 \\
WebArena      & \textcolor{agentblue}{\textbf{.270}}/.860/.667 & .240/\textcolor{uiorange}{\textbf{.980}}/.000 & .210/.800/\textcolor{envteal}{\textbf{.778}} \\
TravelGym     & \textcolor{agentblue}{\textbf{1.245}}/.230/.845 & 1.165/\textcolor{uiorange}{\textbf{.280}}/\textcolor{envteal}{\textbf{.930}} & 1.186/.125/.789 \\
SWE-bench     & .019/\textcolor{uiorange}{\textbf{.040}}/\textcolor{envteal}{\textbf{.050}} & .016/.004/.040 & \textcolor{agentblue}{\textbf{.028}}/.013/.023 \\
\midrule
Win count     & 4/2/2/0 & 1/1/1/0 & 0/1/1/2 \\
\bottomrule
\end{tabular}%
}
\caption{Amateur persona across benchmarks (productivity / proactivity / personalization, means). Best per column in each row: \textcolor{agentblue}{productivity} (blue), \textcolor{uiorange}{proactivity} (orange), \textcolor{envteal}{personalization} (teal). Bottom row: number of model combos (tau2/WA/TG/SWE) where that mode was best or tied for best in productivity.}
\label{tab:case-amateur}
\end{table}

Table~\ref{tab:case-amateur} breaks down amateur-persona productivity across the four benchmarks. $M_{\text{text}}$ is the best or tied-for-best mode in 4 of 4 $\tau^2$-bench combos, 2 of 2 WebArena combos, and 2 of 4 TravelGym combos. Only SWE-bench deviates, where all modes score near zero and $M_{\text{hybrid}}$ holds a marginal edge (.028 vs.\ .019). The explanation is consistent with the persona design: amateur users refuse complex questions and reward only very simple inquiries. $M_{\text{ui}}$ and $M_{\text{hybrid}}$ encourage richer interaction, but against an amateur persona, the extra communication turns yield mostly ``I don't know'' responses and accumulate cost without recovering useful information. $M_{\text{text}}$, by asking fewer and simpler questions, avoids wasted turns.

\smallskip
\noindent\textbf{Case 2: one\_question persona. $M_{\text{ui}}$ excels at personalization but sacrifices productivity.}

\begin{table}[htbp]
\centering
\resizebox{\columnwidth}{!}{%
\begin{tabular}{lccc}
\toprule
\textbf{Benchmark} & \textbf{$M_{\text{text}}$} & \textbf{$M_{\text{ui}}$} & \textbf{$M_{\text{hybrid}}$} \\
\midrule
$\tau^2$-bench & .315/.080/.395 & \textcolor{agentblue}{\textbf{.325}}/\textcolor{uiorange}{\textbf{.105}}/\textcolor{envteal}{\textbf{.760}} & \textcolor{agentblue}{\textbf{.325}}/.070/.275 \\
WebArena      & \textcolor{agentblue}{\textbf{.260}}/.717/.364 & .180/\textcolor{uiorange}{\textbf{.910}}/\textcolor{envteal}{\textbf{1.000}} & .240/.600/.242 \\
TravelGym     & \textcolor{agentblue}{\textbf{1.186}}/.110/.790 & 1.122/\textcolor{uiorange}{\textbf{.245}}/\textcolor{envteal}{\textbf{.950}} & 1.167/.098/.711 \\
SWE-bench     & .122/\textcolor{uiorange}{\textbf{.076}}/.601 & .103/.062/\textcolor{envteal}{\textbf{.733}} & \textcolor{agentblue}{\textbf{.140}}/.040/.681 \\
\bottomrule
\end{tabular}%
}
\caption{one\_question persona across benchmarks (productivity / proactivity / personalization). Best per column in each row: \textcolor{agentblue}{productivity} (blue), \textcolor{uiorange}{proactivity} (orange), \textcolor{envteal}{personalization} (teal).}
\label{tab:case-oneq}
\end{table}

Table~\ref{tab:case-oneq} shows a consistent trade-off for one\_question users: $M_{\text{ui}}$ achieves the highest personalization in every benchmark (by margins of 0.16--0.76 over $M_{\text{hybrid}}$), but trails in productivity in three of four benchmarks. The persona's interaction preference (answer exactly one question per turn) maps naturally onto UI-based communication, where the agent can present a single focused selection widget. However, the structured UI interaction is slower than asking one targeted text question, leading to lower throughput. This persona illustrates the \emph{personalization--productivity tension}: the mode that best satisfies the user's interaction style is not always the mode that completes the task fastest.

\section{CPE-Optimized Policies}
\label{sec:cpe-policies}

Below are the best communication policy patches discovered by CPE for each benchmark and model pairing. For SWE-bench, the patch rewrites the full hybrid system prompt and examples; for $\tau^2$-bench, TravelGym, and WebArena, it appends a suffix to the agent's system prompt. We reproduce the suffix and example portions.

\subsection{SWE-bench}

\begin{promptbox}[agentblue]{SWE-bench: DeepSeek-V3.2 + Qwen3-VL-32B (hybrid example)}
{\small
--------------------- EXAMPLES (hybrid: choose text or UI per turn) ---------------------

For each clarification, pick \textbf{one} tool: \textbf{ask\_question} (user reads plain text in chat) or \textbf{generate\_ui} (user sees a rendered HTML page). Use \textbf{ask\_question} when a short sentence is enough; use \textbf{generate\_ui} when layout, choices, or structured fields help.

--- Example A --- text (\texttt{ask\_question}) after inspecting the repo ---

USER: The app sometimes shows a random error page. Can you figure out what's wrong?

ASSISTANT: I'll list the top-level layout, then ask one focused text question if details are missing. \\
\texttt{execute\_bash: ls -1 /testbed}

USER: api / frontend / scripts / tests

ASSISTANT: \texttt{ask\_question: What is the exact error message or stack trace when the error page appears?}

--- Example B --- visual (\texttt{generate\_ui}) when structured input helps ---

ASSISTANT (needs options or a clearer form): \\
\texttt{generate\_ui:} \\
\texttt{progress\_summary: I've mapped the repo. I need one concrete detail to reproduce the issue.} \\
\texttt{html\_code:} A minimal HTML form with a \texttt{<select>} for timing and a \texttt{<textarea>} for additional notes, styled with sans-serif font.
}
\end{promptbox}

\subsection{$\tau^2$-bench}

\begin{promptbox}[envteal]{$\tau^2$-bench: DeepSeek-V3.2 + Qwen3-VL-32B}
When a user provides their user ID, immediately use it to look up their account and reservations without asking for it again. For users who prefer one question at a time, ensure each \texttt{ask\_question} contains only a single, clear query. For users who can only answer selection questions, always phrase questions with explicit lettered options (A, B, C, etc.) and use \texttt{ask\_question}. Use \texttt{generate\_ui} only when presenting complex visual choices or forms, not for simple selection or yes/no questions. For amateur users, avoid asking for specific dates, times, or policy details; ask only general, non-technical questions. If a user's constraints (e.g., `do not transfer') are known, respect them and avoid suggesting options that violate them.
\end{promptbox}

\begin{promptbox}[envteal]{$\tau^2$-bench: DeepSeek-V3.2 + GPT-4o}
Clarification policy (append-only): Prefer \texttt{ask\_question} for single short factual checks (reservation/user/payment ID, yes/no, a single-letter or A/B/C choice) and whenever the user explicitly prefers one-question-at-a-time. Use \texttt{generate\_ui} when you need structured input, when presenting more than three detailed options, or when a visual layout/form will make selection or entry unambiguous --- always include a 1--3 sentence progress\_summary and exactly one question object in the UI. Ask only the minimum data required to take the next action and verify identity once at the start of the session (don't repeatedly re-request the same credential unless the session expired). If an \texttt{ask\_question} is answered ``I don't know'' twice or the UI times out, switch strategy: either present a short \texttt{generate\_ui} (single question) or rephrase as a concise A/B/C \texttt{ask\_question} ($\leq$3 choices). Never repeat an identical question; if clarification fails, offer an alternate distinguishing question or a short selection UI. One function call per turn; keep clarifications focused, short, and directly tied to the next system action.
\end{promptbox}

\begin{promptbox}[envteal]{$\tau^2$-bench: GPT-5-mini + GPT-4o}
Allow at most 2 consecutive \texttt{ask\_question} probes for the same verification goal; if the user replies ``I don't know'' twice in a row about that item, stop repeating low-yield probes and immediately present a single escalation step (one short \texttt{ask\_question}) offering three clear choices (e.g.\ A) Transfer to human, B) Send OTP to contact on file, C) Search by passenger name+date). Use \texttt{ask\_question} only for single short answers or a single-letter/number selection and honor pref\_one\_question/pref\_do\_selection by asking one concise selection; use \texttt{generate\_ui} only when you need the user to enter 3+ structured fields, sensitive exact values, or the user explicitly agrees to complete a form. When calling \texttt{generate\_ui} include a 1--2 sentence progress\_summary, render exactly one question in the questions JSON that matches the HTML, and include only the fields strictly necessary.
\end{promptbox}

\begin{promptbox}[envteal]{$\tau^2$-bench: GPT-5-mini + Qwen3-VL-32B}
Attempt \texttt{get\_reservation\_details} or \texttt{get\_user\_details} before asking anything; if all required fields are present, call the action immediately. Use \texttt{ask\_question} only for exactly one missing fact or a single-choice decision (prefer A/B/C); keep the prompt concise ($\leq$20 words), wrap it as \texttt{<parameter=query>...</parameter>}, and ask exactly one question per call. Limit consecutive \texttt{ask\_question} calls to 2; if the user replies ``I don't know'' or times out twice, stop asking and call \texttt{transfer\_to\_human\_agents} with a one-sentence summary. Use \texttt{generate\_ui} only when you must collect two-or-more structured fields and the user is likely able to complete a small form; include a 1--3 sentence progress\_summary, a complete HTML5 document, and a questions parameter that exactly matches the page question.
\end{promptbox}

\subsection{TravelGym}

\begin{promptbox}[uiorange]{TravelGym: DeepSeek-V3.2 + GPT-4o}
UI/question policy: When the user is vague, prefer one focused database search (\texttt{interact\_with\_env choice=search}) to gather candidates before asking clarifying questions. Use \texttt{interact\_with\_env} with \texttt{choice=action} and \texttt{content\_kind=text} for a single, concise clarification (ask one thing only). Use \texttt{content\_kind=html} only to present a small selection UI (roughly 2--6 labeled options) --- the HTML must be a complete valid HTML5 document, include \texttt{<meta charset="utf-8"/>}, and present exactly one clear travel-related question. After the user answers, if you can confidently recommend, call \texttt{interact\_with\_env} with \texttt{choice=answer} and return exactly one option ID. Conserve interaction rounds, keep questions short and targeted, match the user's stated amateur/pro tone, and avoid unnecessary proactive suggestions.
\end{promptbox}

\begin{promptbox}[uiorange]{TravelGym: GPT-5-mini + Qwen3-VL-32B}
Hybrid interaction rules: Use \texttt{interact\_with\_env choice=action} when you need to ask the user something. For short clarifications (dates, budget, party size, location, one yes/no or single preference) use \texttt{content\_kind=text} and ask exactly one focused question per action. Use \texttt{content\_kind=html} only to present a structured selection (at most 3 clearly labeled options) --- supply a complete valid HTML5 document with \texttt{<meta charset="utf-8"/>} and make the page ask exactly one travel-related question. Use \texttt{choice=search} to query the database once you have the necessary constraints. Use \texttt{choice=answer} only to return a single option ID after preferences are confirmed. Avoid multi-part compound questions; ask sequentially to gather required details efficiently.
\end{promptbox}

\subsection{WebArena}

\begin{promptbox}[hybridpurple]{WebArena: DeepSeek-V3.2 + GPT-4o}

\textbf{Clarification Strategy.}
Before using \texttt{ask\_question} or \texttt{generate\_ui}, first explore the page thoroughly using goto, click, scroll, and other navigation actions. Use \texttt{ask\_question} for simple, text-based clarifications about the task (e.g., `Which category should I search in?'). Use \texttt{generate\_ui} only when you need to present multiple options visually or gather structured input that benefits from a form. Keep questions concise and task-focused; avoid asking for information already visible on the page or implied by the task description. If the task is ambiguous after exploration, ask one clarifying question to resolve the ambiguity, then proceed.

\textbf{Action Priority.}
Prioritize direct navigation and interaction over asking questions. Only use \texttt{ask\_question} or \texttt{generate\_ui} when you've exhausted visible page information and cannot proceed without clarification. When you do ask, make your question specific and actionable --- reference what you've already tried or observed on the page.
\end{promptbox}

\begin{promptbox}[hybridpurple]{WebArena: GPT-5-mini + GPT-4o}
Because productivity is the primary objective, prefer completing the task by interacting with visible page controls (\texttt{click}/\texttt{fill}/\texttt{select}/\texttt{scroll}) and by exploring the AXTree/URL before asking the user anything. Use \texttt{ask\_question} only when a single, specific missing fact truly blocks progress and cannot be discovered by on-page actions; phrase it as one short sentence ($\leq$20 words) that names the missing fact and why it is needed, and cite a page element or AXTree id if applicable. Use \texttt{generate\_ui} only when the user must choose among multiple explicit options or complete a small structured form that cannot be captured in one short question; include a 1--2 sentence progress\_summary tied to the current AXTree/URL, keep the html minimal, and include exactly one question. Never ask about information obtainable by simple exploration; avoid repeated clarification turns (one clarifying question per blocking ambiguity).
\end{promptbox}

\subsection{Analysis of Optimized Policies}

Several patterns recur across the CPE-discovered policies, reflecting convergent strategies for effective channel selection:

\begin{enumerate}[nosep]
\item \textbf{Environment-first, then clarify.} SWE-bench and WebArena policies instruct the agent to explore the repository or page before asking questions. TravelGym policies recommend a database search before questioning. This avoids wasting communication turns on information the agent can discover autonomously.

\item \textbf{Text for simple, UI for structured.} All policies converge on a consistent channel heuristic: \texttt{ask\_question} for single factual queries (IDs, yes/no, A/B/C choices), \texttt{generate\_ui} for structured input (3+ fields, visual layouts, option comparisons). This aligns with the low/high-constraint design of the two primitives.

\item \textbf{Fail-fast escalation.} $\tau^2$-bench policies independently discover a ``2-strikes'' rule: after two consecutive ``I don't know'' or timeout responses, stop probing and escalate (transfer to human, offer structured alternatives). This prevents unproductive clarification loops.

\item \textbf{Persona-aware adaptation.} The $\tau^2$-bench DeepSeek-V3.2 + Qwen3-VL-32B policy explicitly adapts to all four personas (amateur: no professional questions; do\_selection: A/B/C only; one\_question: single query; answer\_more: forms allowed). Other policies encode this implicitly through question constraints.

\end{enumerate}

These patterns emerge purely from productivity-driven prompt optimization without human specification, confirming that CPE discovers effective and interpretable communication heuristics.

\section{CPE Configuration}
\label{sec:cpe-config}

All CPE runs share the same optimization setup, differing only in the candidate dataset and model pairing. We optimize the communication policy prompt (system prompt, examples, or environment-specific suffix); agent and user-simulator weights are frozen. Table~\ref{tab:cpe-config} lists the key parameters. $K$ is the number of episodes evaluated per round; the reflect LLM receives up to 5 trajectory excerpts (only \texttt{ask\_question} / \texttt{generate\_ui} turns) from the $K$ rollouts. $R$ is the maximum number of rounds; in practice, the best policy was found within 25 rounds across all runs.

\begin{table*}[htbp]
\centering
\caption{CPE configuration across benchmarks.}
\label{tab:cpe-config}
\small
\begin{tabular}{lcccc}
\toprule
& \textbf{SWE-bench} & \textbf{TravelGym} & \textbf{$\tau^2$-bench} & \textbf{WebArena} \\
\midrule
Episodes ($|\mathcal{D}|$) & 100 & 200 & 200 & 200 \\
Train fraction & 0.7 & 0.7 & 0.7 & 0.7 \\
Candidates per round ($K$) & 5 & 10--20 & 5--20 & 10 \\
Max rounds ($R$) & 30 & 30 & 30 & 30 \\
Reflect model & \multicolumn{4}{c}{Same as agent model} \\
Trajectories in prompt & \multicolumn{4}{c}{5 per round; only \texttt{ask\_question} / \texttt{generate\_ui} turns shown} \\
\bottomrule
\end{tabular}
\end{table*}

\section{LLM Usage}

LLMs were used only for light editing (grammar and clarity of writing) and minor code completion (data processing scripts). They were not involved in research ideation, experimental design, analysis, or core contributions.

\end{document}